\documentclass[runningheads]{llncs}
\usepackage{graphicx}
\usepackage{amsmath}
\usepackage{amsmath,amssymb,amsfonts}
\usepackage{graphicx}
\usepackage{textcomp}
\usepackage{xcolor}

\usepackage{algorithm}
\usepackage{algorithmic}

\usepackage[T1]{fontenc}

\usepackage{epstopdf}

\DeclareMathOperator{\sigu}{\textsc{sigmoid\_u}}
\DeclareMathOperator{\sigb}{\textsc{sigmoid\_b}}
\DeclareMathOperator{\sine}{\textsc{sine}}
\DeclareMathOperator{\satu}{\textsc{satlin\_u}}
\DeclareMathOperator{\satb}{\textsc{satlin\_b}}
\DeclareMathOperator{\relu}{\textsc{relu}}
\DeclareMathOperator{\soft}{\textsc{softplus}}

\begin{document}
\title{Data-Driven Learning of Feedforward Neural Networks with Different Activation Functions\thanks{Supported by Grant 2017/27/B/ST6/01804 from the National Science Centre, Poland.}}
\titlerunning{Data-Driven Learning of FNN with Different Activation Functions}
\author{Grzegorz Dudek\orcidID{0000-0002-2285-0327}}
\authorrunning{G. Dudek}
\institute{Electrical Engineering Faculty, Częstochowa University of Technology,\\ Częstochowa, Poland\\
\email{grzegorz.dudek@pcz.pl}}
\maketitle              % typeset the header of the contribution
\begin{abstract}

This work contributes to the development of a new data-driven method (D-DM) of feedforward neural networks (FNNs) learning. This method was proposed recently as a way of improving randomized learning of FNNs by adjusting the network parameters to the target function fluctuations. The method employs logistic sigmoid activation functions for hidden nodes. In this study, we introduce other activation functions, such as bipolar sigmoid, sine function, saturating linear functions, reLU, and softplus. We derive formulas for their parameters, i.e. weights and biases. In the simulation study, we evaluate the performance of FNN data-driven learning with different activation functions. The results indicate that the sigmoid activation functions perform much better than others in the approximation of complex, fluctuated target functions.

\end{abstract}

\section{Introduction}

FNNs are widely used as predictive models to fit data distribution. They learn using gradient descent methods and ensure a universal approximation property. However, gradient-based algorithms suffer from many drawbacks which make the learning process ineffective and time-consuming. This is because gradient learning is sensitive to local minima, flat regions, and saddle points of the loss function. Moreover, its application is time-consuming for complex target functions (TFs), big data, and large FNN architectures. Randomized learning was proposed as an alternative to gradient-based learning. In this approach, the parameters of the hidden nodes are selected randomly from any interval, and stay fixed. Only the output weights are learned. The optimization problem in randomized learning becomes convex and can be solved by a standard linear least-squares method \cite{Pri15}. This leads to very fast training. The the universal approximation property is kept when the random parameters are selected from a symmetric interval according to any continuous sampling distribution \cite{Hus99}. The main problems in randomized learning are \cite{Cao18}, \cite{Zha16}: how to select the interval and distribution for the random parameters, and whether the weights and biases should be chosen from the same interval and distribution.

It was shown in \cite{Dud19} and \cite{Dud20a} that the weights and biases of hidden nodes have different functions and should be selected separately. The weights decide about the activation function (AF) slopes and should reflect the TF complexity, while the biases decide about the AF shift and should ensure the placement of the most nonlinear fragments of AFs into the input hypercube. These fragments are most useful for modeling TF fluctuations. The method proposed in \cite{Dud19} selects the proper interval for the weights based on AF features and TF properties. The biases are calculated based on the weights and data scope. This approach introduces the AFs into the input hypercube and adjusts the interval for weights to TF complexity. In \cite{Dud20a}, instead of generating the weights, the slope angles of AFs were randomly selected. This changed the distribution of weights, which typically is a uniform one. This new distribution ensured that the slope angles of AFs were uniformly distributed, which improved results by preventing overfitting, especially for highly nonlinear TFs.

To improve further FNN randomized learning, in \cite{Dud20}, a D-DM was proposed. This method introduces the AFs into randomly selected regions of the input space and adjusts the AF slopes to the TF slopes in these regions. As a result, the AFs mimic the TF locally, and their linear combination approximates smoothly the entire TF. This work contributes to the development of data-driven FNN learning by introducing different AFs, i.e. bipolar sigmoid, sine function, saturating linear functions, reLU, and softplus. For each AF, the formulas for weights and biases are derived.   

The remainder of this paper is structured as follows. In Section 2, the framework of D-DM is presented. The formulas for hidden nodes parameters for different AFs are derived in Section 3. The performance of FNN data-driven learning with different AFs is evaluated in Section 4. Finally, Section 5 concludes the work.

%Sine_AF_Parascandolo17.pdf

%https://math.stackexchange.com/questions/1469739/neural-network-differentiate-bipolar-sigmoidal-function

%https://wandb.ai/shweta/Activation%20Functions/reports/A-Comparative-Study-of-Activation-Functions--VmlldzoxMDQwOTQ

\section{Framework of the Data-Driven FNN Learning}

Let us consider a shallow FNN architecture with $n$ inputs, a single-hidden layer, and a single output. AFs of hidden nodes, $h(\mathbf{x})$, map nonlinearly input vectors $\mathbf{x}=[x_1, x_2,..., x_n]^T \in \mathbb{R}^n$ into an $m$-dimensional feature space. An output node combines linearly $m$ nonlinear transformations of the inputs. The function expressed by this FNN has the form:   

\begin{equation}
\varphi(\mathbf{x}) = \sum_{i=1}^{m}\beta_ih_i(\mathbf{x})
\label{eq1}
\end{equation}
where $\beta_i$ is the output weight linking the $i$-th hidden node with the output node.

Such FNN architecture has a universal approximation property, even when the hidden layer parameters are not trained but generated randomly from the proper distribution \cite{Igel95}, \cite{Hus99}. 

The output weights $ \boldsymbol{\beta} = [\beta_1, \beta_2, ..., \beta_m]^T$ can be determined by solving the following linear problem: $\mathbf{H}\boldsymbol{\beta} = \mathbf{Y}$, where $\mathbf{H} = [\mathbf{h}(\mathbf{x}_1), \mathbf{h}(\mathbf{x}_2),  ..., \mathbf{h}(\mathbf{x}_N)]^T \in \mathbb{R}^{N \times m}$ is the hidden layer output matrix, and $ \mathbf{Y} = [y_1, y_2,  ..., y_N]^T $ is a~vector of target outputs. The optimal solution for $ \boldsymbol{\beta}$ is given by:

\begin{equation}
\boldsymbol{\beta} = \mathbf{H}^+\mathbf{Y}
\label{eq2}
\end{equation}
where $ \mathbf{H}^+ $ denotes the Moore–Penrose generalized inverse of matrix $ \mathbf{H} $.

The hidden node parameters, i.e. weights $ \mathbf{a} = [ a_{1}, a_{2}, ..., a_{n}]^T$ and a bias $b$, control slopes and position of AF in the input space. For a sigmoid AF given by the formula: 

\begin{equation}
h(\mathbf{x}) = \frac{1}{1 + \exp\left(-\left(\mathbf{a}^T\mathbf{x} + b\right)\right)}
\label{eq3}
\end{equation}
weight $a_j$ decides about the sigmoid slope in the $j$-th direction and bias $b$ decides about the sigmoid shift along a hyperplane containing all $x$-axes. The appropriate selection of the slopes and shifts of all sigmoids determine the fitting accuracy of FNN to the TF. To adjust the sigmoids to the local features of the TF, in \cite{Dud20}, a D-DM for FNN learning was proposed. This method selects an input space region by randomly choosing one of the training points for each sigmoid. Then, it places the sigmoid in this region and adjusts the sigmoid slopes to the TF slopes in the neighborhood of the chosen point. By combining linearly all the sigmoids randomly placed in the input space, we obtain a fitted surface which reflects the TF shape in different regions. 

The D-DM algorithm, in the first step, selects randomly training point $\mathbf{x}^*$. Then, sigmoid $S$ is placed in the input space in such a way that one of its inflection points, $P$, is in $\mathbf{x}^*$. The sigmoid value at the inflection point is $0.5$:  

\begin{equation}
h(\mathbf{x}^*) = \frac{1}{1 + \exp\left(-\left(\mathbf{a}^T\mathbf{x}^* + b\right)\right)}=0.5
\label{eq4}
\end{equation}

From this equation we obtain the sigmoid bias as:

\begin{equation}
b = -\mathbf{a}^T\mathbf{x}^*
\label{eq5}
\end{equation}

The slopes of sigmoid $S$ are adjusted to the TF slopes in $\mathbf{x}^*$. The TF slopes in $\mathbf{x}^*$ are estimated by fitting hyperplane $T$ to the neighborhood of $\mathbf{x}^*$. The neighborhood, $\Psi(\mathbf{x}^*)$, contains point $\mathbf{x}^*$ and $k$ training points nearest to it. Hyperplane $T$ has the form:

\begin{equation}
y = a_1'x_1 + a_2'x_2+...+a_n'x_n +b'
\label{eq6}
\end{equation}
where coefficient $a_j'$ expresses a slope of $T$ in the $j$-th direction.  

We assume that sigmoid $S$ is tangent to hyperplane $T$ in point $\mathbf{x}^*$. This means that the partial derivatives of $S$ and $T$ in $\mathbf{x}^*$ are the same. Comparing the formulas for partial derivatives of both functions, we obtain an equation for the sigmoid weights (see \cite{Dud20} for details):

\begin{equation}
a_j = 4a_j', \quad  j = 1, 2, ..., n
\label{eq7}
\end{equation}   

To generate all the FNN hidden nodes, the D-DM algorithm repeats the procedure described above $m$ times. So, for each node it randomly selects training point $\mathbf{x}^*$, fits hyperplane $T$ to its neighborhood $\Psi(\mathbf{x}^*)$, calculates weights $a_j$ according to \eqref{eq7}, and calculates biases $b$ according to \eqref{eq5}. Finally, it calculates hidden layer output matrix $\mathbf{H}$, and output weights from \eqref{eq2}. The resulting function, $\varphi(\mathbf{x})$, constructed in line with such data-driven learning, reflects TF fluctuations.

The D-DM has two hyperparameters: the number of hidden nodes $m$ and neighbourhood size $k$. They 
control the fitting performance of the model and its bias-variance tradeoff. Their optimal values for a given TF should be tuned during cross-validation.  

\section{Data-Driven FNN Learning with Different Activation Functions}

When we employ other AFs instead of logistic sigmoids, the projection matrix $\mathbf{H}$ changes in a way which can entail changes in the approximation properties of the model. Using other AFs requires the derivation of new formulas for the hidden node parameters in the following ways.

\begin{description}
\item[Bipolar sigmoid] $\sigb$. 
Usually the bipolar sigmoid is defined as a hyperbolic tangent function. In this study, we define it slightly differently:

\begin{equation}
h_{sigb}(\mathbf{x}) = \frac{2}{1 + \exp\left(-\left(\mathbf{a}^T\mathbf{x} + b\right)\right)} -1
\label{eq8}
\end{equation}

D-DM places $\sigb$ in the input space in such a way that one of its inflection points is in the randomly selected training point, $\mathbf{x}^*$. The $\sigb$ value at the inflection points is $0$, so, $h_{sigb}(\mathbf{x}^*)=0$. From this equation we obtain the formula for the bias, which is the same as for the unipolar sigmoid ($\sigu$), \eqref{eq5}.

To find weights $a_j$, we equate the partial derivatives of $\sigb$ in $\mathbf{x}^*$ to the partial derivatives of hyperplane $T$, \eqref{eq6}:
%https://math.stackexchange.com/questions/1469739/neural-network-differentiate-bipolar-sigmoidal-function

\begin{equation} 
\frac{\partial h_{sigb}(\mathbf{x}^*)}{\partial x_j} = \frac{1}{2}a_j(1+h_{sigb}(\mathbf{x}^*))(1-h_{sigb}(\mathbf{x}^*)) = a_j'
\label{eq9}
\end{equation}

From this equation, taking into account that $h_{sigb}(\mathbf{x}^*)=0$, we obtain:

\begin{equation}
a_j = 2a_j', \quad  j = 1, 2, ..., n
\label{eq10}
\end{equation} 

\item[Sine function] $\sine$. 
%https://stats.stackexchange.com/questions/402618/can-sinx-be-used-as-activation-in-deep-learning
Let us place the $\sine$ AF, $h_{sin}(\mathbf{x})=\sin(\mathbf{a}^T\mathbf{x} + b)$, in the input space in such a way that it has one of its inflection point in randomly selected training point $\mathbf{x}^*$. The $\sine$ value in the inflection points is 0, so, $h_{sin}(\mathbf{x}^*)=0$. From this equation we obtain the formula for bias, which is the same as for both sigmoid AFs, \eqref{eq5}.  

To determine equations for the weights for $\sine$, we equate the partial derivatives of $\sine$ in $\mathbf{x}^*$ to the partial derivatives of hyperplane $T$, \eqref{eq6}:

\begin{equation} 
\frac{\partial h_{sin}(\mathbf{x}^*)}{\partial x_j} = a_j\cos(\mathbf{a}^T\mathbf{x}^* + b) = a_j'
\label{eq11}
\end{equation}

Taking into account that $\sin(\mathbf{a}^T\mathbf{x}^* + b)=0$ implies $\cos(\mathbf{a}^T\mathbf{x}^* + b)=1$, from \eqref{eq11} we obtain:

\begin{equation}
a_j = a_j', \quad  j = 1, 2, ..., n
\label{eq12}
\end{equation} 

\item[Saturating linear unipolar function] $\satu$.
This is a linearized version of $\sigu$ defined as follows:

\begin{equation}
	h_{satu}(\mathbf{x}) =
	\left\lbrace 
	%\[\arraycolsep=4.4pt\def\arraystretch{1}
	\begin{array}{llll}
		0 & & \mathrm{if} & z \leq 0 \\
		z & & \mathrm{if} & 0 < z < 1\\
		1 & & \mathrm{if} & z \geq 1
	\end{array}
	\right.
	\label{eq13}
\end{equation}
where $z=\mathbf{a}^T\mathbf{x} + b$. 

$\satu$ is placed in the input space in such a way that it has a value of 0.5 in $\mathbf{x}^*$. This is analogous to $\sigu$ to which $\satu$ has a similar shape. Thus, $\mathbf{a}^T\mathbf{x}^* + b=0.5$. From this equation we obtain:

\begin{equation}
b = 0.5 -\mathbf{a}^T\mathbf{x}^*
\label{eq14}
\end{equation}

We assume that the middle segment of $h_{satu}(\mathbf{x})$, $\mathbf{a}^T\mathbf{x}+b$, has the same slopes as hyperplane $T$, thus:

\begin{equation}
a_j = a_j', \quad  j = 1, 2, ..., n
\label{eq15}
\end{equation} 

\item[Saturating linear bipolar function] $\satb$.
This AF is a linearized version of bipolar sigmoid $\sigb$: 

\begin{equation}
	h_{satb}(\mathbf{x}) =
	\left\lbrace 
%	\[\arraycolsep=4.4pt\def\arraystretch{1}
	\begin{array}{llll}
		-1 & & \mathrm{if} & z \leq -1 \\
		z & &\mathrm{if} & -1 < z < 1\\
		1 & &\mathrm{if} & z \geq 1
	\end{array}
	\right.
	\label{eq16}
\end{equation}
where $z=\mathbf{a}^T\mathbf{x} + b$. 

$\satb$ is placed in the input space in such a way that it has a value of 0 in $\mathbf{x}^*$. Thus, $\mathbf{a}^T\mathbf{x}^* + b=0$. From this equation we obtain the same formula for a bias as for sigmoid AFs, \eqref{eq5}.

As with $\satu$, we assume that the middle segment of $\satb$ has the same slopes as hyperplane $T$. Thus, weights $a_j$ are the same as the $T$ coefficients, \eqref{eq15}.

\item[Rectified linear unit] $\relu$.
This is an AF commonly used in deep learning. It is expressed by:

\begin{equation}
	h_{reLU}(\mathbf{x}) =
	\left\lbrace 
	%\[\arraycolsep=4.4pt\def\arraystretch{1}
	\begin{array}{llll}
		0 & & \mathrm{if} & z \leq 0 \\
		z & & \mathrm{if} & z > 0
	\end{array}
	\right.
	\label{eq17}
\end{equation}
where $z=\mathbf{a}^T\mathbf{x} + b$. 

$\relu$ is composed of two half-hyperplanes: the first being $y = 0$ and the second $y=\mathbf{a}^T\mathbf{x} + b$. D-DM places the $\relu$ AF in the input space so that the second half-hyperplane coincides with hyperplane $T$. Thus, their coefficients are the same:

\begin{equation}
b = b', \quad a_j = a_j', \quad  j = 1, 2, ..., n
\label{eq18}
\end{equation} 

\item[Softplus] $\soft$.
This is a smooth approximation of the $\relu$. It is expressed by:

\begin{equation}
h_{soft}(\mathbf{x}) = \ln \left(1 + \exp\left(\mathbf{a}^T\mathbf{x} + b\right)\right)
\label{eq19}
\end{equation} 

For $\mathbf{x}=[0,0, ..., 0]$ and $b=0$, the value of $h_{soft}(\mathbf{x})=\ln(2)$. Let us shift this function in such a way that it has the value of $\ln(2)$ in $\mathbf{x}^*$. In such a case $\ln (1 + \exp(\mathbf{a}^T\mathbf{x}^* + b))=\ln(2)$. From this equation we obtain a formula for $b$, which is the same as for the sigmoids \eqref{eq5}.

Now, let us assume that the slopes of $\soft$ in $\mathbf{x}^*$ are the same as the slopes of $T$. Equating the partial derivative of both functions we obtain:

\begin{equation} 
\frac{\partial h_{soft}(\mathbf{x}^*)}{\partial x_j} = \frac{a_j}{1+\exp(-(\mathbf{a}^T\mathbf{x}^* + b))} = a_j'
\label{eq20}
\end{equation}

From $\ln (1 + \exp(\mathbf{a}^T\mathbf{x}^* + b))=\ln(2)$ we obtain $1+\exp(\mathbf{a}^T\mathbf{x}^* + b)=2$. Substituting this into \eqref{eq20}, we obtain the weights of hidden nodes with $\soft$ AFs:

\begin{equation}
a_j = 2a_j', \quad  j = 1, 2, ..., n
\label{eq21}
\end{equation} 

\end{description}

Table \ref{tab1} details the hidden nodes parameters determined by D-DM for different AFs. Note that in all cases, weights $a_j$ reflect hyperplane $T$ coefficients $a_j'$. Biases for all AFs, excluding $\relu$, are expressed using a dot product of the weight vector and $\mathbf{x}^*$ vector.    

\begin{table}[h]
\caption{Hidden nodes parameters for different activation functions.}
\label{tab1}
\begin{center}
\setlength{\tabcolsep}{7pt}
%\begin{tabular}{@{}rlcc{}}
\begin{tabular}{rlcc}
\hline
& Activation function  & Weights $a_j$ & Bias $b$ \\ \hline
$\sigu$: &	$h_{sigu}(\mathbf{x}) = \frac{1}{1 + \exp\left(-\left(\mathbf{a}^T\mathbf{x} + b\right)\right)}$ & $4a_j'$ &	$-\mathbf{a}^T\mathbf{x}^*$	\\
$\sigb$: &	$h_{sigb}(\mathbf{x}) = \frac{2}{1 + \exp\left(-\left(\mathbf{a}^T\mathbf{x} + b\right)\right)} -1$ &	$2a_j'$ & $-\mathbf{a}^T\mathbf{x}^*$		\\ 
$\sine$: &	$h_{sin}(\mathbf{x})=\sin(\mathbf{a}^T\mathbf{x} + b)$ &	$a_j'$ & $-\mathbf{a}^T\mathbf{x}^*$	\\
$\satu$: &	$h_{satu}(\mathbf{x}) =
	\left\lbrace 
	\begin{array}{llll}
		0 & & \mathrm{if} & z \leq 0 \\
		z & & \mathrm{if} & 0 < z < 1\\
		1 & & \mathrm{if} & z \geq 1
	\end{array}
	\right.$ & $a_j'$ & $0.5 -\mathbf{a}^T\mathbf{x}^*$ 	\\
$\satb$: &	$h_{satb}(\mathbf{x}) =
	\left\lbrace 
	\begin{array}{llll}
		-1 & & \mathrm{if} & z \leq -1 \\
		z & &\mathrm{if} & -1 < z < 1\\
		1 & &\mathrm{if} & z \geq 1
	\end{array}
	\right.$ & $a_j'$ & $-\mathbf{a}^T\mathbf{x}^*$ 	\\
	
$\relu$: &	$h_{reLU}(\mathbf{x}) =
	\left\lbrace 
	%\[\arraycolsep=4.4pt\def\arraystretch{1}
	\begin{array}{llll}
		0 & & \mathrm{if} & z \leq 0 \\
		z & & \mathrm{if} & z > 0
	\end{array}
	\right.$ & $a_j'$ & $b'$ 	\\
$\soft$: &	$h_{soft}(\mathbf{x}) = \ln \left(1 + \exp\left(\mathbf{a}^T\mathbf{x} + b\right)\right)$ &	$2a_j'$ & $-\mathbf{a}^T\mathbf{x}^*$		\\ 
\hline
\end{tabular}

	{\raggedright 
	where $a_j'$ and $b'$ are coefficients of hyperplane $T$, $y = a_1'x_1 + a_2'x_2+...+a_n'x_n +b'$, adjusted to the TF in the neighborhood $\Psi(\mathbf{x}^*)$ of randomly selected training point $\mathbf{x}^*$; $z=\mathbf{a}^T\mathbf{x} + b$. \par}
\end{center}
\end{table}

\begin{figure}[h]
\centering
\includegraphics[width=0.32\textwidth]{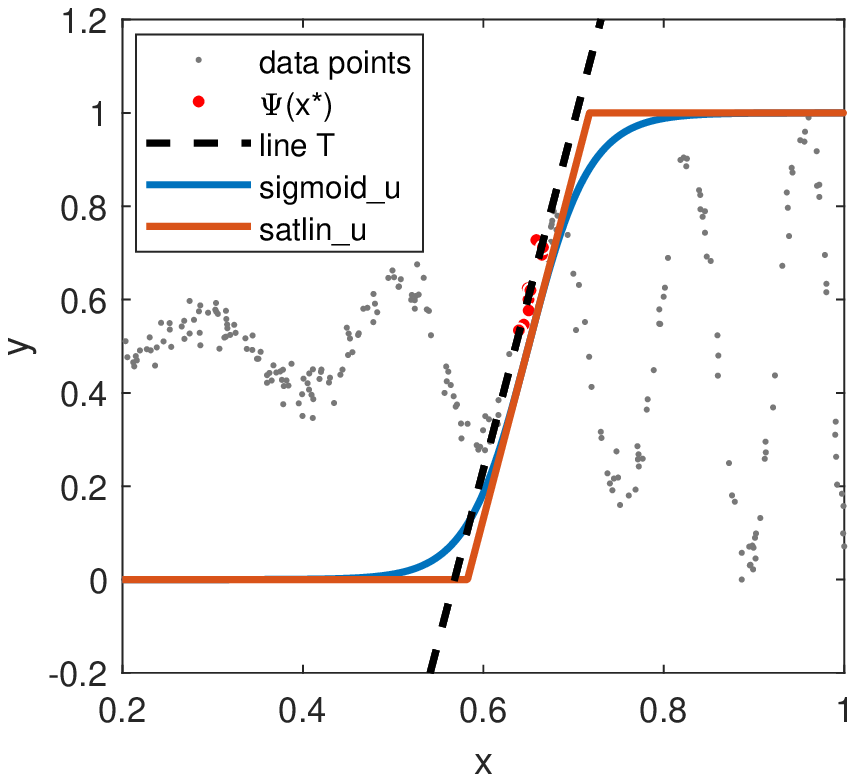}
\includegraphics[width=0.32\textwidth]{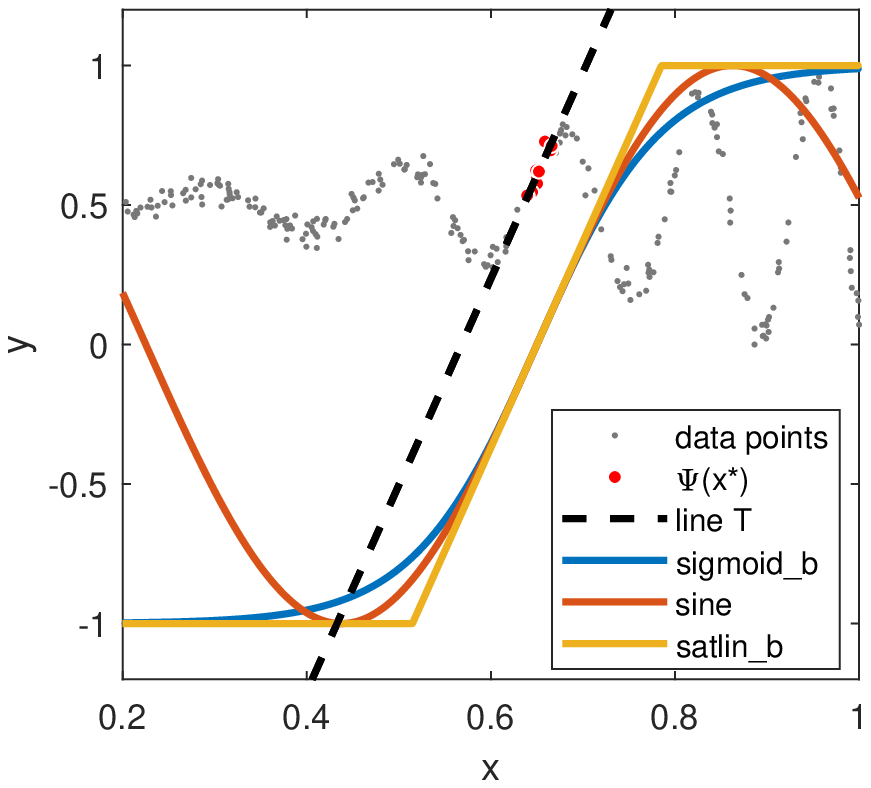}
\includegraphics[width=0.32\textwidth]{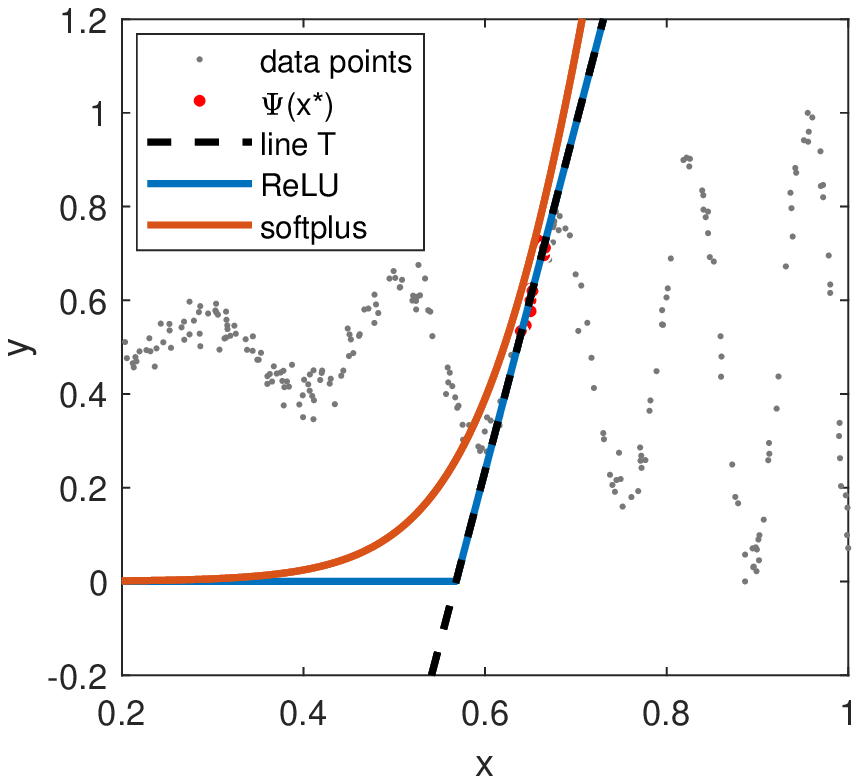}
\caption{AFs of different types introduced into the input space in $\mathbf{x}^*$ by D-DM.} \label{figSc}
\end{figure}

Fig. \ref{figSc} shows AFs of different types introduced into the input space by D-DM. The training points belonging to the neighborhood of $\mathbf{x}^*$, $\Psi(\mathbf{x}^*)$, are shown as red dots. Note that the AFs in all cases have the same slopes in $\mathbf{x}^*$ as the slope of line $T$, which estimates the TF slope in $\mathbf{x}^*$. D-DM introduces $m$ AFs in different regions of the input space.

\section{Simulation Study}

In this section, we report the experimental results over several regression problems in order to compare the fitting properties of D-DM with different AFs, . They include an approximation of extremely nonlinear TFs: 

\begin{description}
\item[TF1] $g(x) = \sin\left(20\cdot\exp x \right)\cdot x^2, \, x \in [0, 1]$ 

\item[TF2] $g(x) = 0.2e^{-\left(10x - 4\right)^2} + 0.5e^{-\left(80x - 40\right)^2} + 0.3e^{-\left(80x - 20\right)^2}, \, x \in [0, 1]$. 

\item[TF3] $g(\mathbf{x}) = \sum_{j=1}^{n}\sin\left(20\cdot\exp x_j\right)\cdot x_j^2, \, x_i \in [0, 1]$ 

\item[TF4] $g(\mathbf{x}) = -{\sum_{i=1}^{n} \sin(x_i) \sin^{20} \left(\frac{ix_i^2}{\pi}   \right)}, \, x_i \in [0, \pi]$ 

\item[TF5] $g(\mathbf{x}) = 418.9829n -{\sum_{i=1}^{n} x_i \sin(\sqrt{|x_i|})}, \, x_i \in [-500, 500]$

\end{description}

\begin{figure}
\centering
\includegraphics[width=0.243\textwidth]{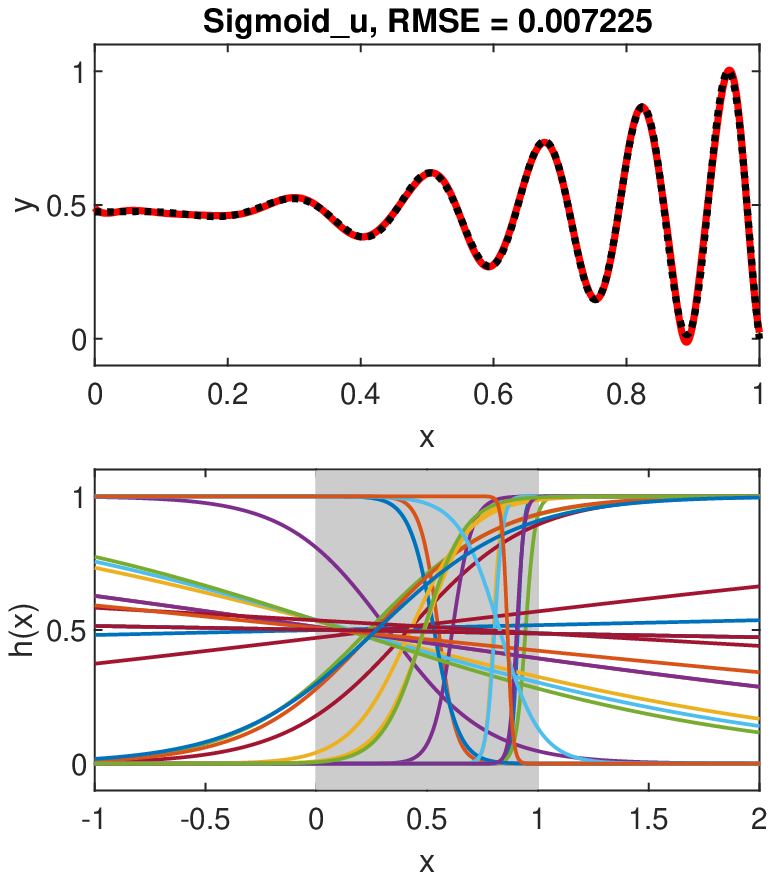}
\includegraphics[width=0.243\textwidth]{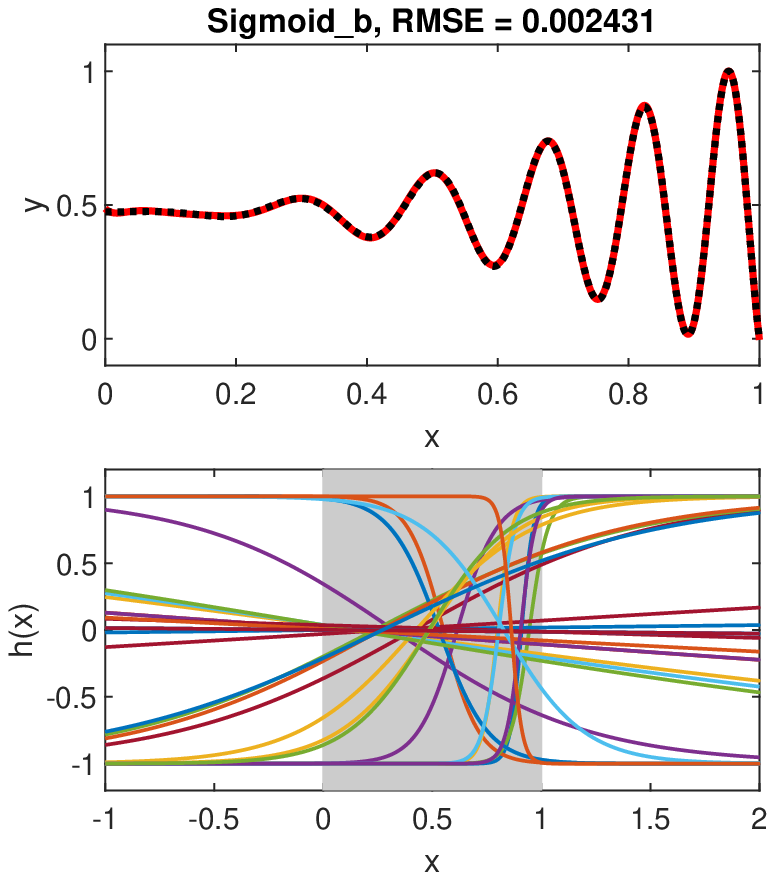}
\includegraphics[width=0.243\textwidth]{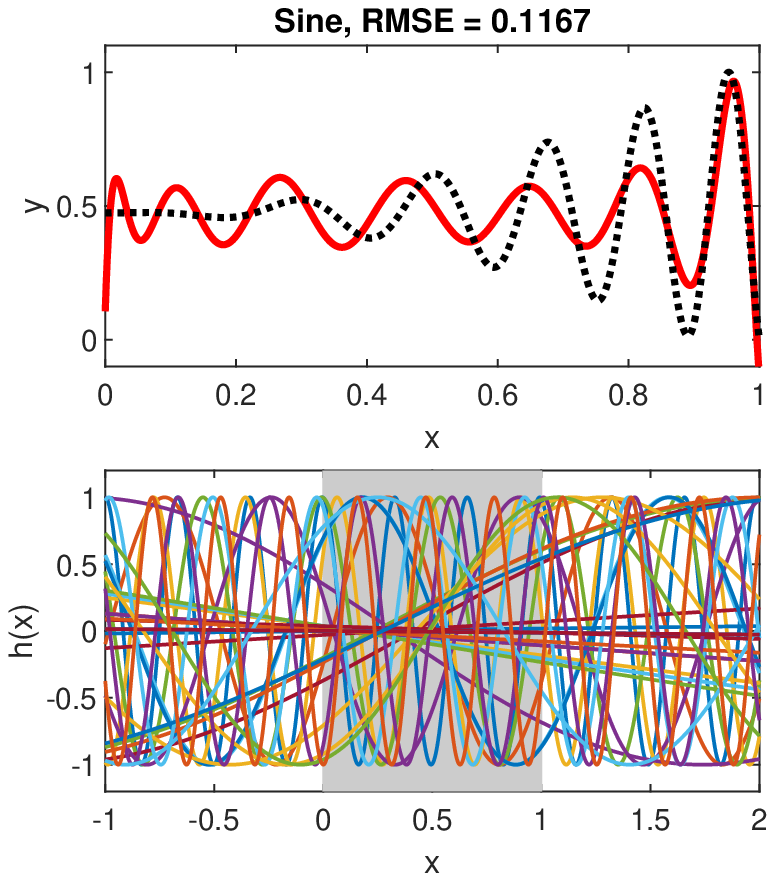}
\includegraphics[width=0.243\textwidth]{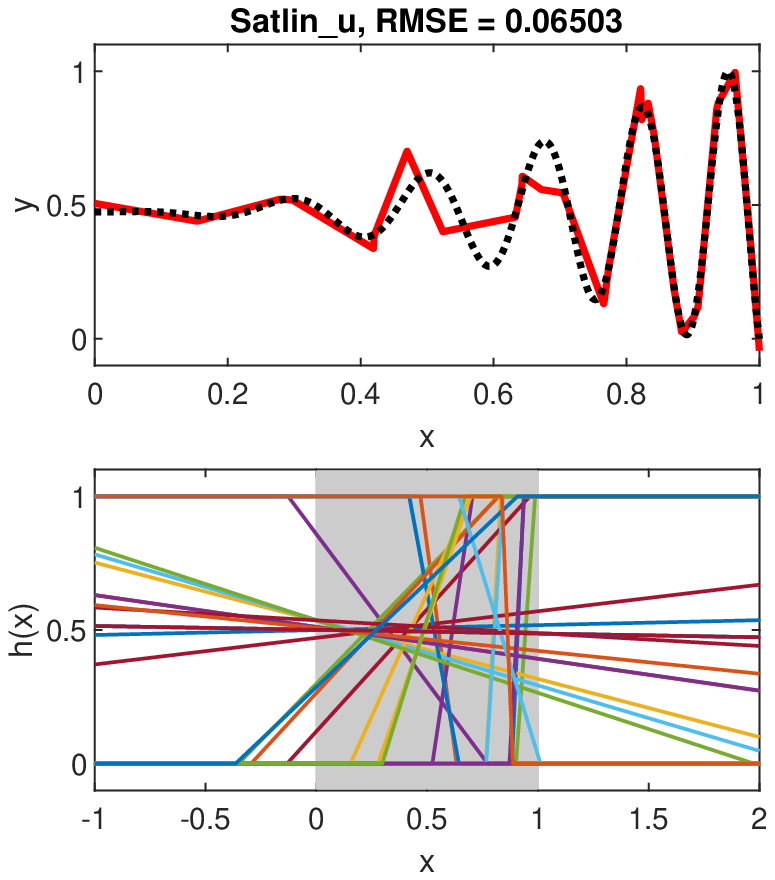}
\includegraphics[width=0.243\textwidth]{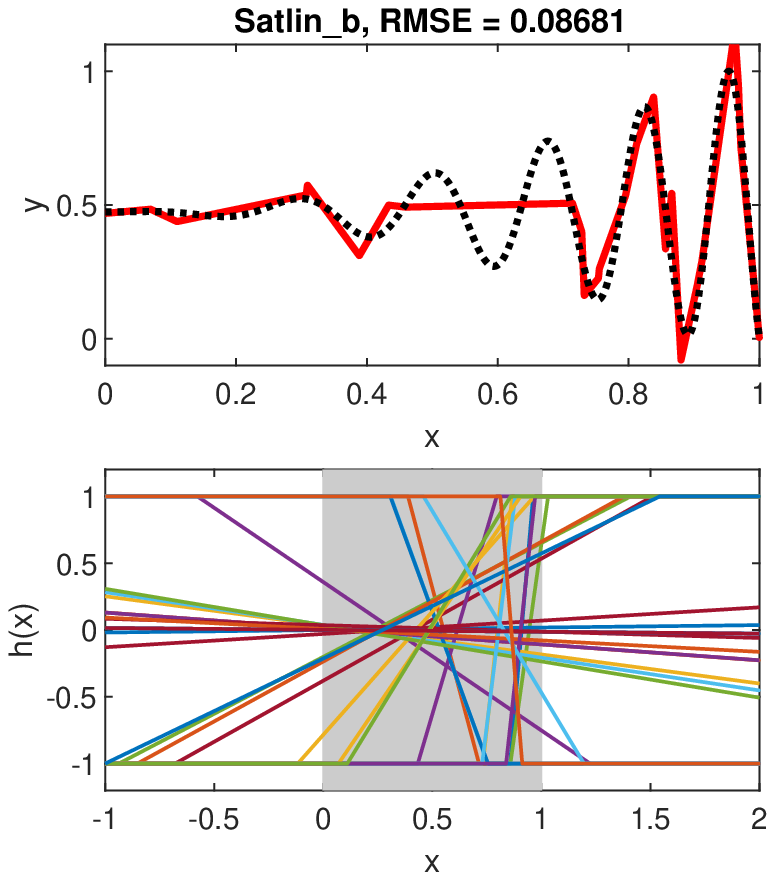}
\includegraphics[width=0.243\textwidth]{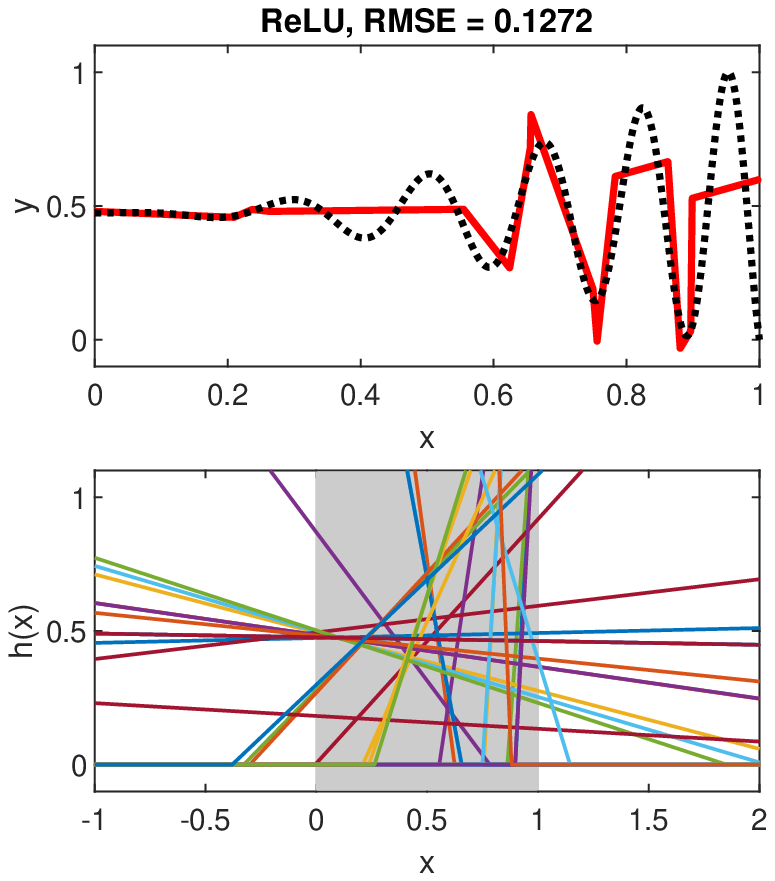}
\includegraphics[width=0.243\textwidth]{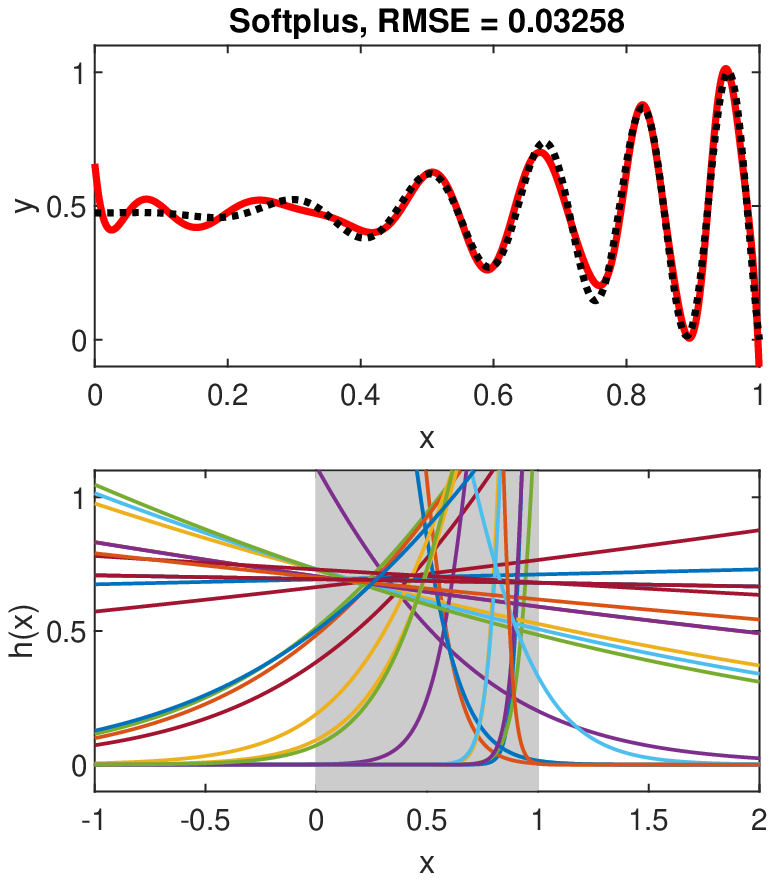}
\includegraphics[width=0.243\textwidth]{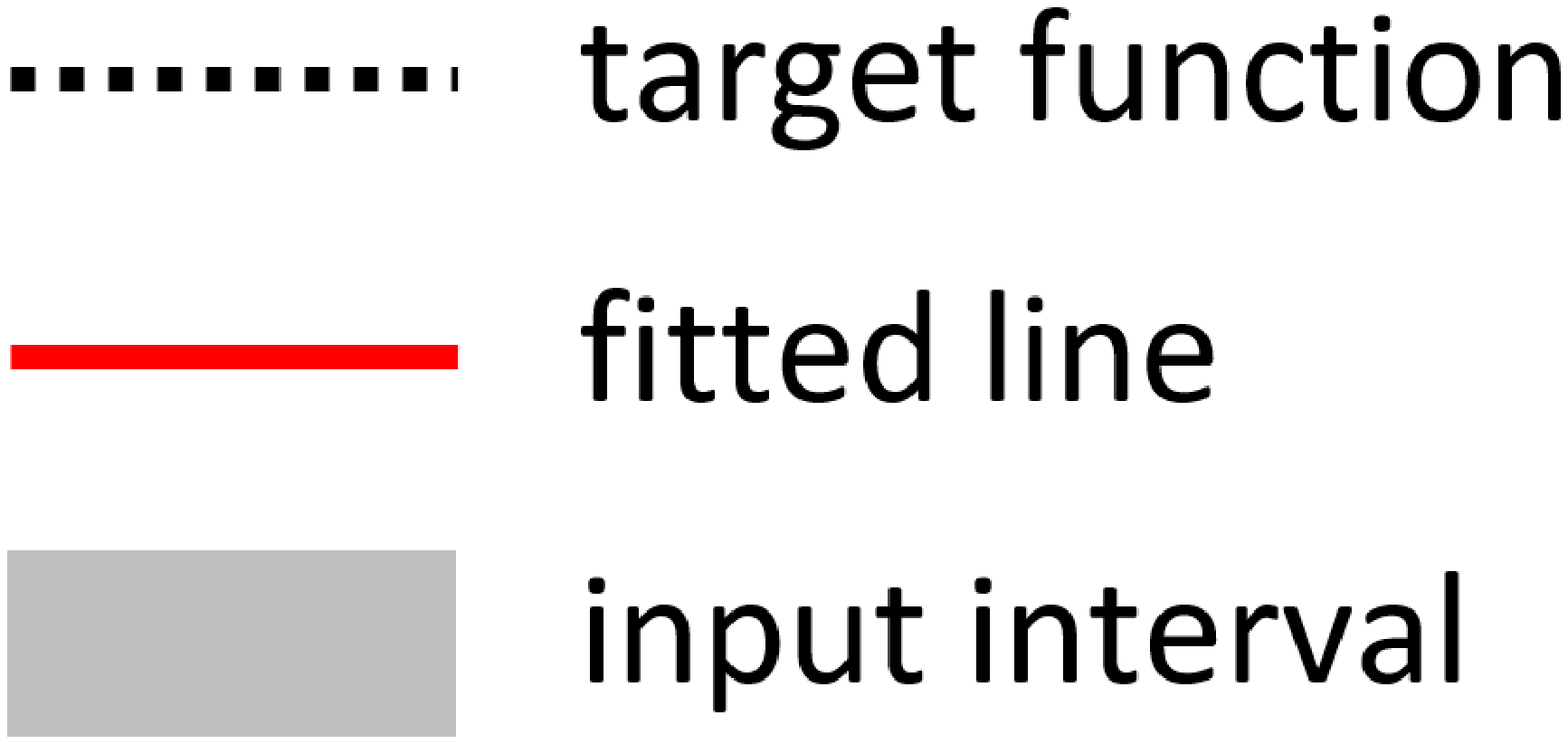}
\caption{TF1: Results of D-DM fitting for different AFs.} \label{figTF1}
\end{figure}

\begin{figure}
%\centering
\includegraphics[width=0.243\textwidth]{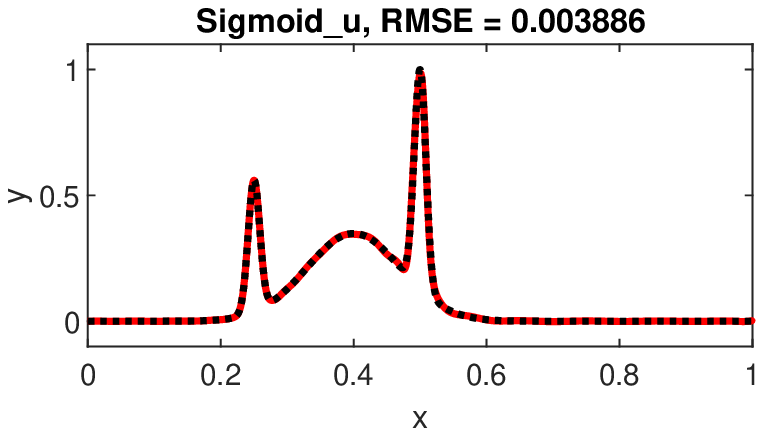}
\includegraphics[width=0.243\textwidth]{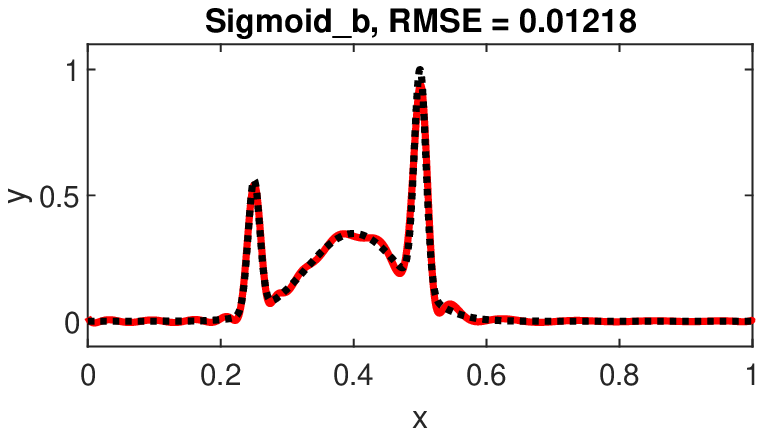}
\includegraphics[width=0.243\textwidth]{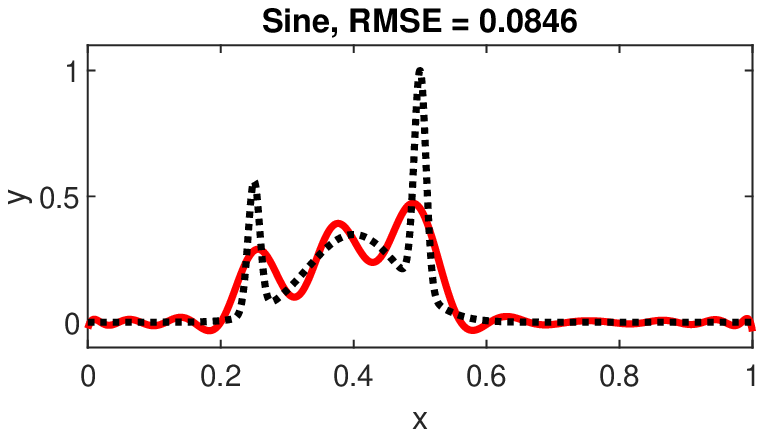}
\includegraphics[width=0.243\textwidth]{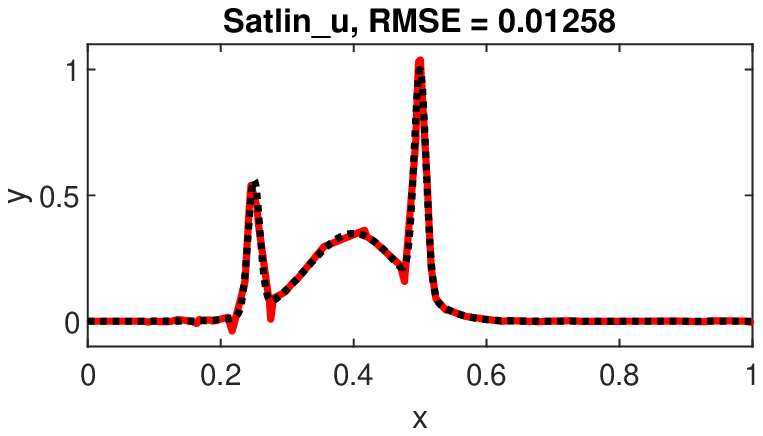}
\includegraphics[width=0.243\textwidth]{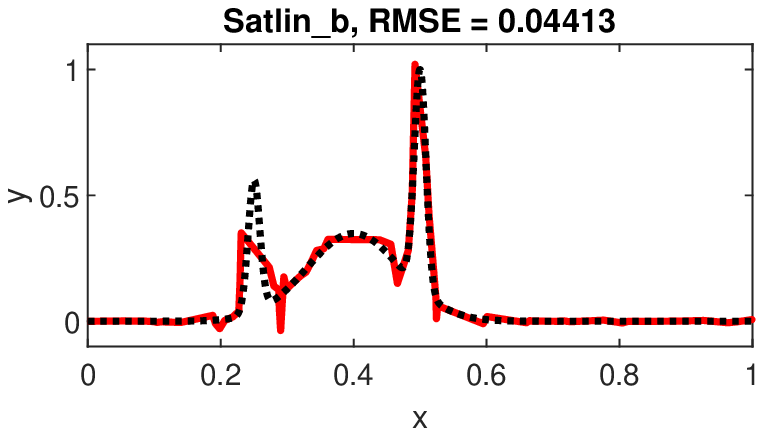}
\includegraphics[width=0.243\textwidth]{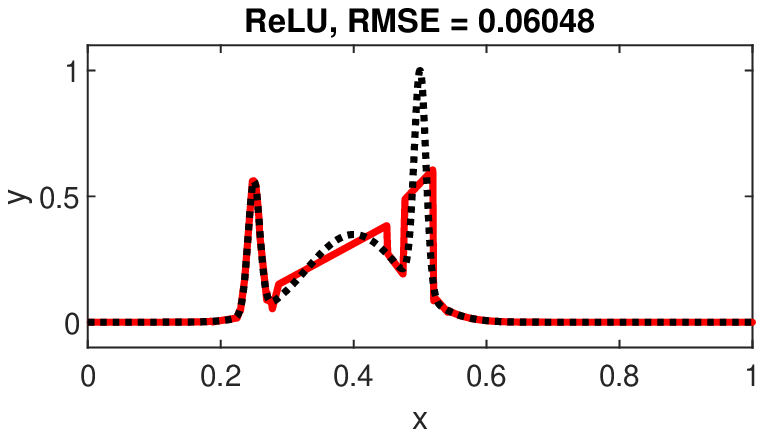}
\includegraphics[width=0.243\textwidth]{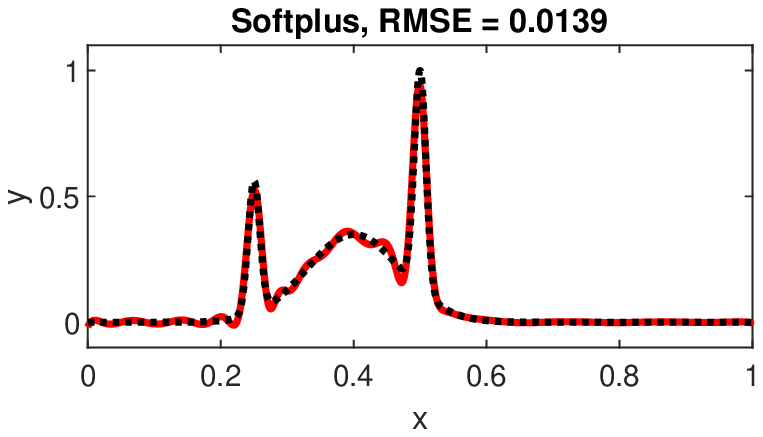}
\includegraphics[width=0.243\textwidth]{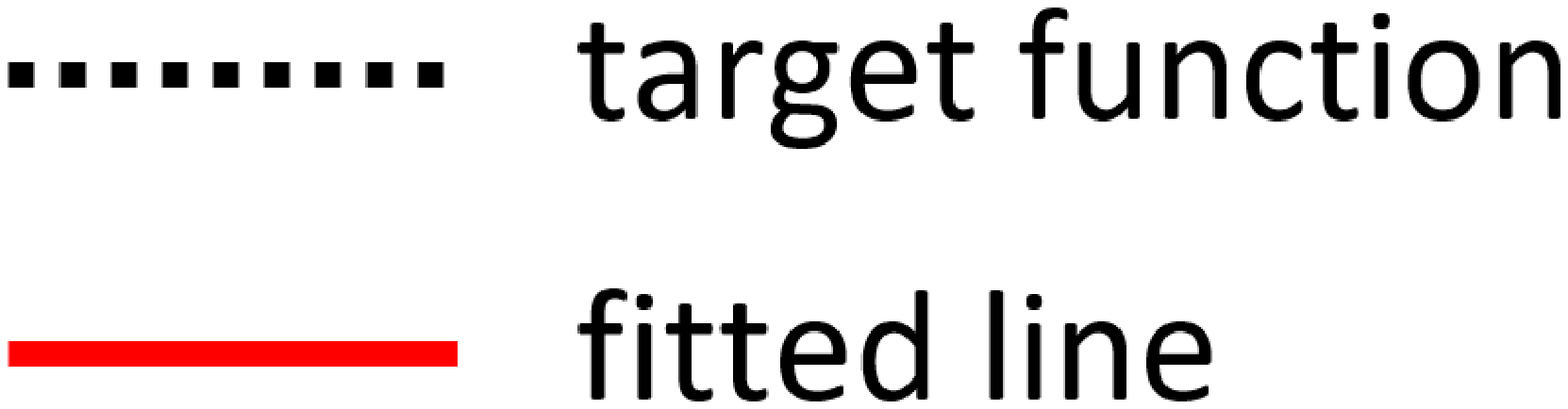}
\caption{TF2: Results of D-DM fitting for different AFs.} \label{figTF2}
\end{figure}

Both the training and test sets for TF1 and TF2 included 5000 points. For the training set, argument $x$ was generated randomly from $U(0,1)$, and for the test set, it was evenly distributed in $[0, 1]$. The function values were normalized in the range $[0, 1]$. Note that TF1 starts flat, near $x = 0$, then has increasing fluctuations (see Fig. \ref{figTF1}). TF2 has two spikes that could be difficult to model with FNN (see Fig. \ref{figTF2}). 

\begin{figure}[]
	%\onecolumn
	\centering
	\includegraphics[width=0.31\textwidth]{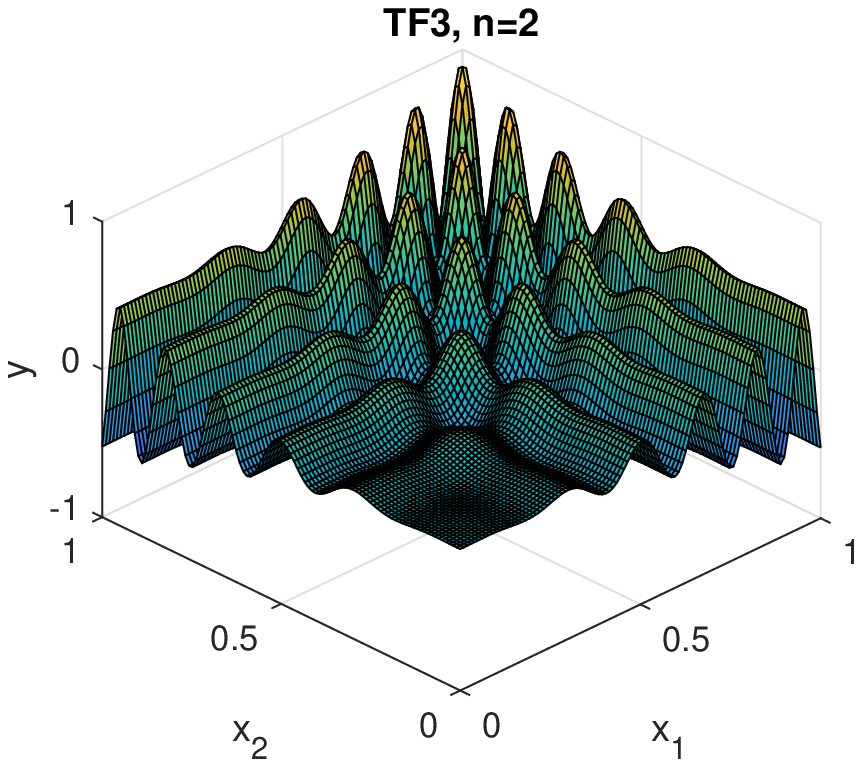}
	\includegraphics[width=0.31\textwidth]{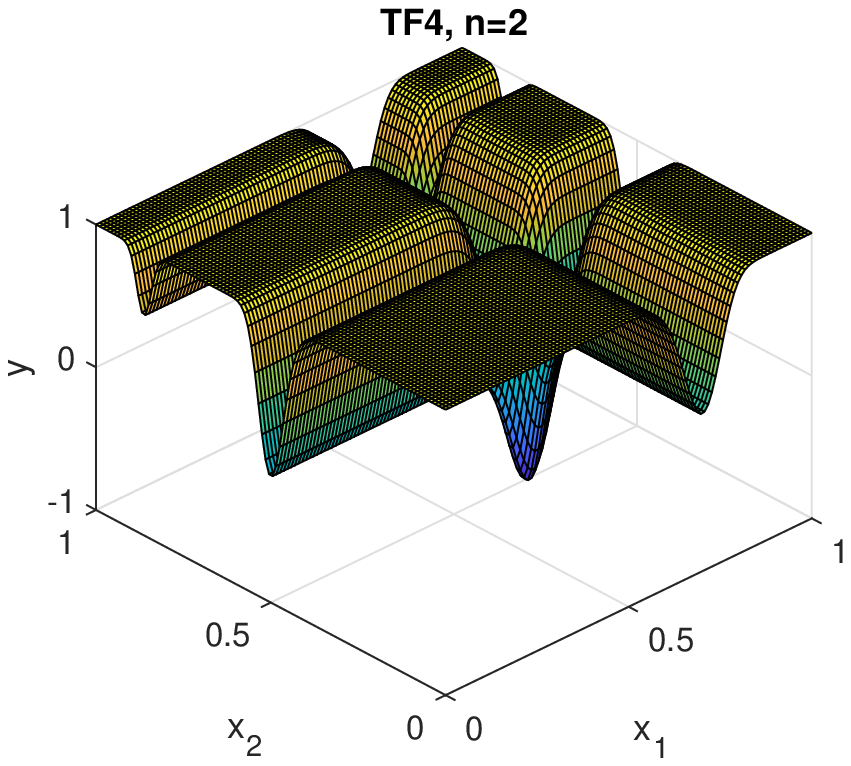}
	\includegraphics[width=0.31\textwidth]{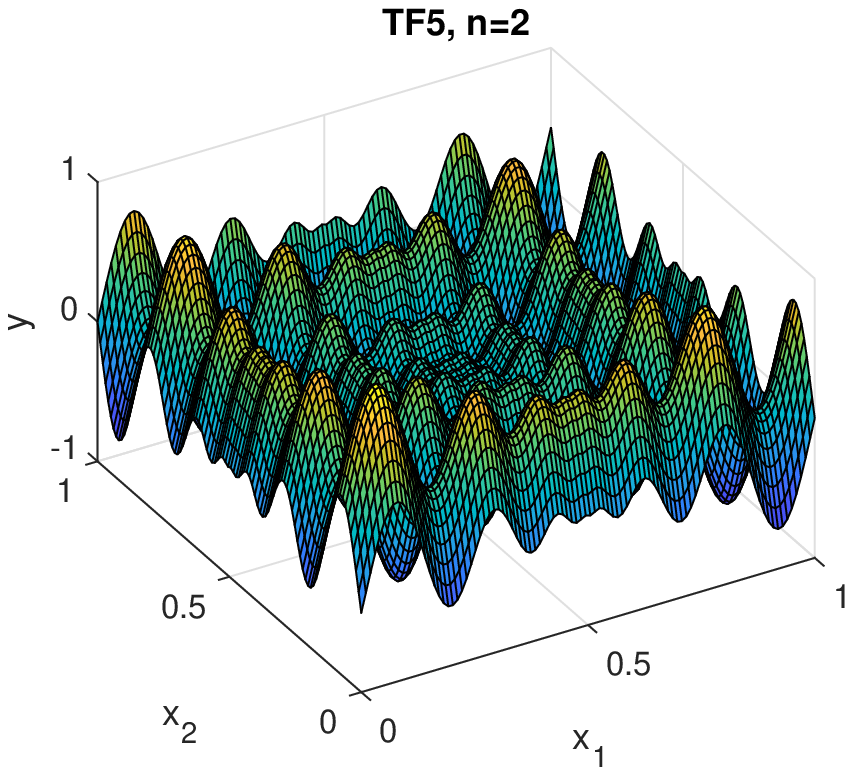}
	\caption{Target functions TF3-TF5.} 
	\label{figTF35}
\end{figure}

TF3-TF5 are multivariate functions. We considered these functions with $n=2, 5$ and 10 arguments. The sizes of the training and test sets depended on the number of arguments. They were 5000 for $n=2$, 20,000 for $n=5$, and 50,000 for $n=10$. All arguments for TF3-TF5 were normalized to $[0, 1]$, and the function values were normalized to $[-1, 1]$. Two-argument functions TF3-TF5 are shown in Fig. \ref{figTF35}. Note that TF3 is a multivariate variant of TF1. It combines flat regions with strongly fluctuated regions. TF4 expresses flat regions with perpendicular grooves. TF5 fluctuates strongly, showing the greatest amplitude at the borders.      

\begin{figure}
\centering
\includegraphics[width=0.49\textwidth]{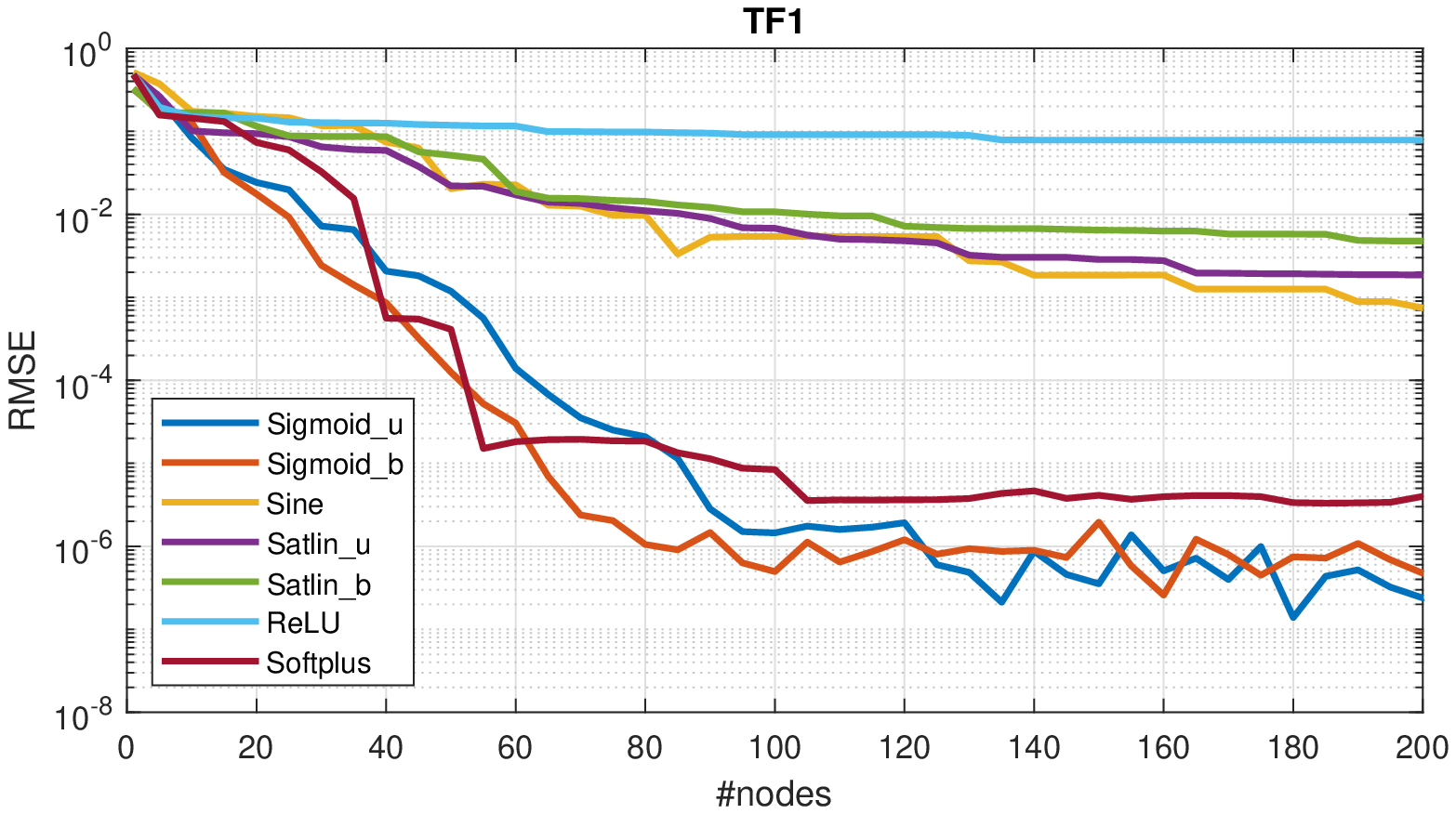}
\includegraphics[width=0.49\textwidth]{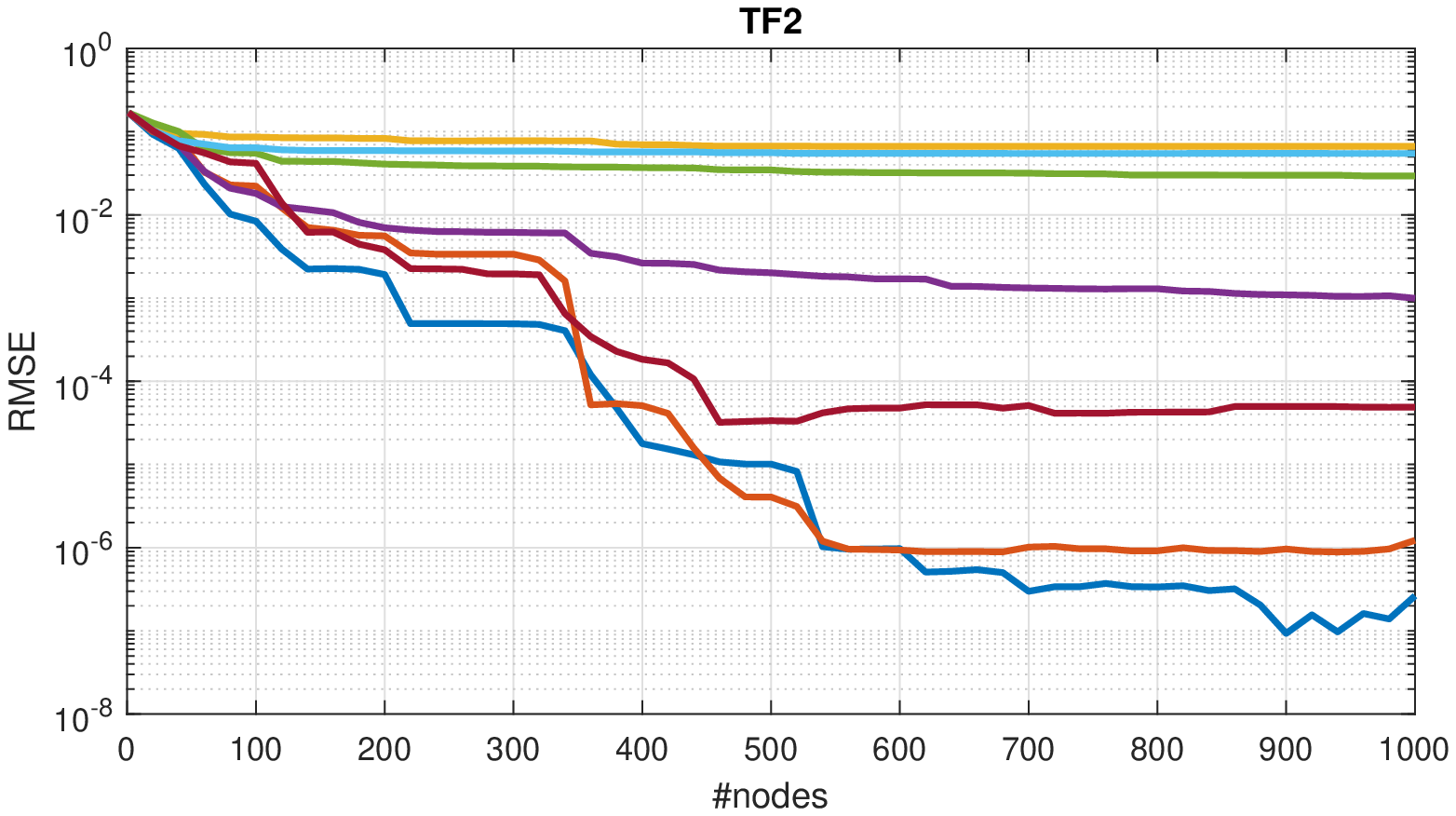}
\caption{Convergence of FNN for TF1 and TF2.} \label{figZb12}
\end{figure}

Fig. \ref{figTF1} shows the results of TF1 fitting. The fitted lines are composed of AFs of different shapes. The AFs distributed by D-DM in the input interval (shown by the gray field) are shown in the lower panels. 
FNN included 30 hidden nodes. The neighborhood size was 2 ($k=1$).
As you can see from Fig. \ref{figTF1}, the slopes of the AFs reflect the TF slopes. D-DM introduces the steepest fragments of the AFs into the input interval. These fragments are the most useful for modeling the TF fluctuations. The saturated AF fragments in the input interval are avoided. The best fitting results were achieved for both sigmoid AFs. $\sine$ cannot cope with a TF with variable intensity of fluctuations. Neither $\relu$, which yielded the highest fitting error, nor the saturating linear functions are not able to fit smoothly to TF1.
The smooth counterpart of $\relu$, $\soft$, improves significantly on the $\relu$ fitting results by offering a smooth approximation of TF1. Obviously, the results are dependent on the number of hidden nodes. The left panel of Fig. $\ref{figZb12}$ shows the TF1 fitting error for different numbers of hidden nodes. As can be seen from this figure, the sigmoid AFs outperformed all the others. Slightly worse results were achieved for $\soft$, while the highest error was observed for $\relu$. Detailed results for each AF, i.e. RMSE for the maximal number of hidden nodes shown in the figures, are presented in Table \ref{tab2}. The lowest errors, i.e. those that are at least 5\% lower than the others, are marked in bold in this table. 

 %$m=30$, $k=2$

 %k=2$

\begin{table}[]
\caption{Fitting errors (RMSE).}
\label{tab2}
\begin{center}
\setlength{\tabcolsep}{2pt}
\begin{tabular}{llrrrrrrr}
\hline
    &        & \multicolumn{1}{c}{$\sigu$} & \multicolumn{1}{c}{$\sigb$} & \multicolumn{1}{c}{$\sine$} & \multicolumn{1}{c}{$\satu$} & \multicolumn{1}{c}{$\satb$} & \multicolumn{1}{c}{$\relu$} & \multicolumn{1}{c}{$\soft$} \\ \hline
TF1 &        & \textbf{2.39E-7}                    & 4.74E-7                    & 7.44E-4                    & 1.86E-3                    & 4.78E-3                    & 7.84E-2                    & 4.00E-6                    \\ 
TF2 &        & \textbf{2.63E-7}                    & 1.23E-6                    & 6.65E-2                    & 9.93E-4                    & 2.93E-2                    & 5.46E-2                    & 4.89E-5                    \\ 
TF3 & $n=2$  & 2.19E-5                    & \textbf{2.26E-6}                    & 1.64E-3                    & 5.81E-3                    & 9.01E-3                    & 1.87E-2                    & -                           \\
TF3 & $n=5$  & 0.2214                      & 0.2215                      & 0.2213                      & 0.2214                      & 0.2215                      & 0.2212                      & -                           \\
TF3 & $n=10$ & 0.2329                      & 0.2328                      & 0.2331                      & 0.2329                      & 0.2328                      & 0.2329                      & -                           \\ 
TF4 & $n=2$  & \textbf{6.69E-7}                    & 4.87E-6                    & 3.95E-2                    & 2.65E-3                    & 9.05E-3                    & 5.18E-2                    & -                           \\
TF4 & $n=5$  & 0.2419                      & 0.2412                      & 0.2411                      & 0.2381                      & 0.2433                      & 0.2418                      & -                           \\
TF4 & $n=10$ & 0.2611                      & 0.2723                      & 0.3095                      & 0.2618                      & 0.2738                      & 0.2571                      & -                           \\ 
TF5 & $n=2$  & \textbf{0.0083}                      & 0.0116                      & 0.0426                      & 0.0257                      & 0.0258                      & 0.0319                      & -                           \\
TF5 & $n=5$  & 0.2385                      & 0.2380                      & 0.2404                      & 0.2390                      & 0.2381                      & 0.2405                      & -                           \\
TF5  & $n=10$ & 0.2246                      & 0.2243                      & 0.2260                      & 0.2247                      & 0.2243                      & 0.2238                      & - \\
    \hline
\end{tabular}
\end{center}
\end{table}

Fig. \ref{figTF2} shows fitting results for TF2 (120 hidden nodes and $k=1$). In this case, $\sigu$ and $\sigb$ provided the best fitting, while $\satu$ and $\soft$ provided a slightly worse fitting. Other AFs could not cope with this TF. For them, increasing the number of hidden nodes did not improve results and RMSE remained outside the acceptable level of 0.01 (see right panel of Fig. $\ref{figZb12}$ and Table \ref{tab2}).

 %$m=120$, $k=2$

\begin{figure}
\centering
\includegraphics[width=0.32\textwidth]{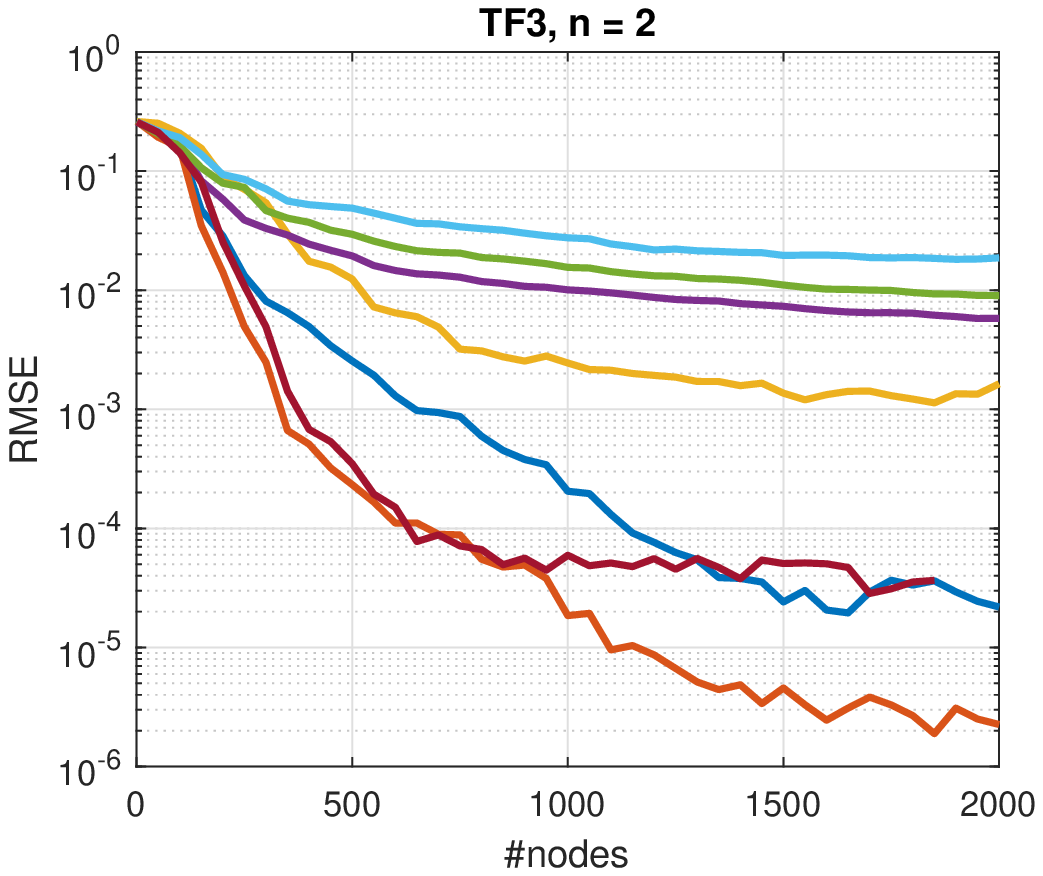}
\includegraphics[width=0.32\textwidth]{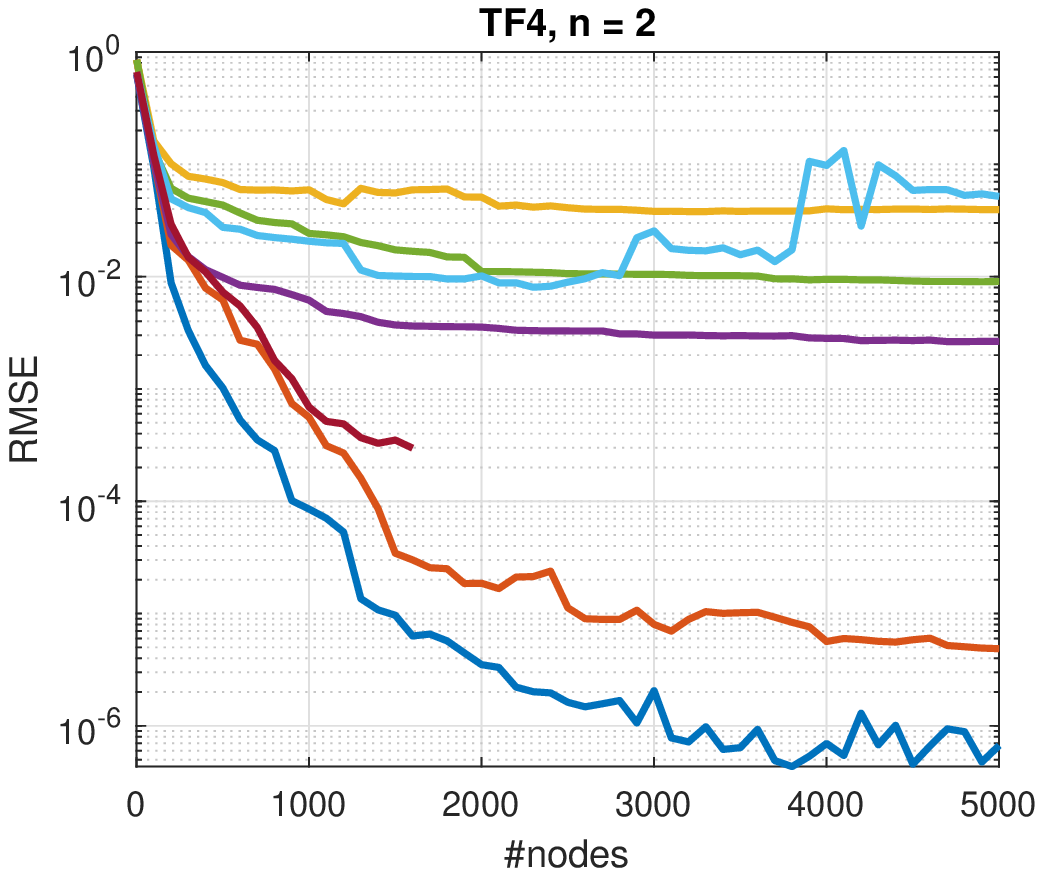}
\includegraphics[width=0.32\textwidth]{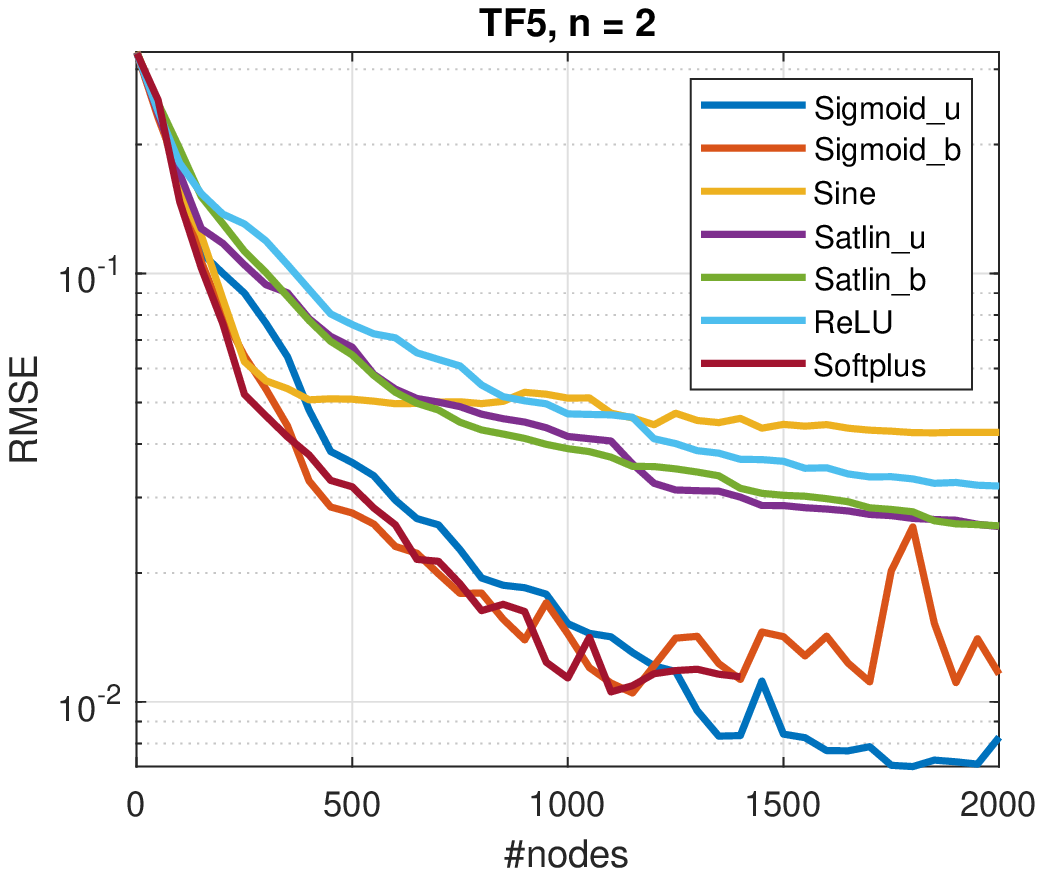}
\caption{Convergence of FNN for TF3-TF5, $n=2$.} \label{figZbn2}
\end{figure}

Fig. \ref{figZbn2} shows the convergence curves of FNN trained using D-DM for two-argument TF3-TF5 ($k=n$). In all these cases, the sigmoid AFs yielded the best results, while $\relu$, $\sine$ and both saturating linear functions yielded the worst results. $\soft$ suffered from numerical problems related to the rapid growth of this function and exceeding the limit for double precision numbers. So, in Table \ref{tab2}, no results for $\soft$ are given. 

\begin{figure}
\centering
\includegraphics[width=0.32\textwidth]{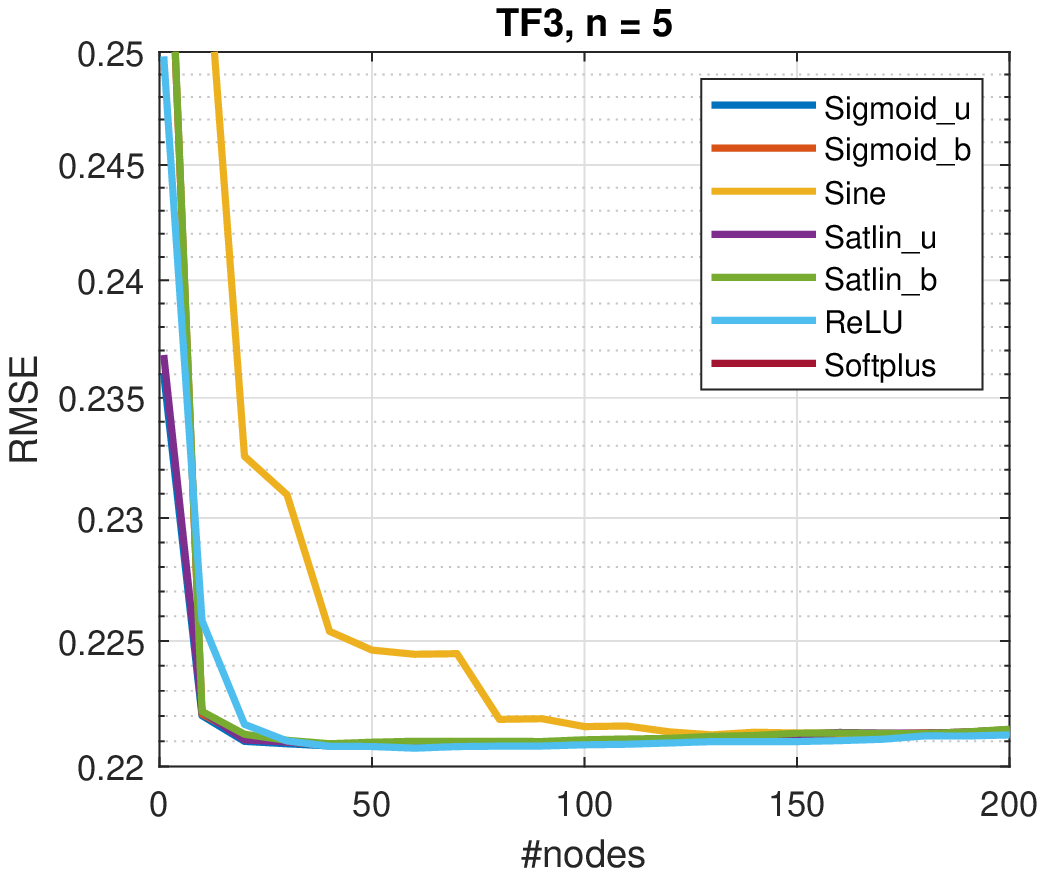}
\includegraphics[width=0.32\textwidth]{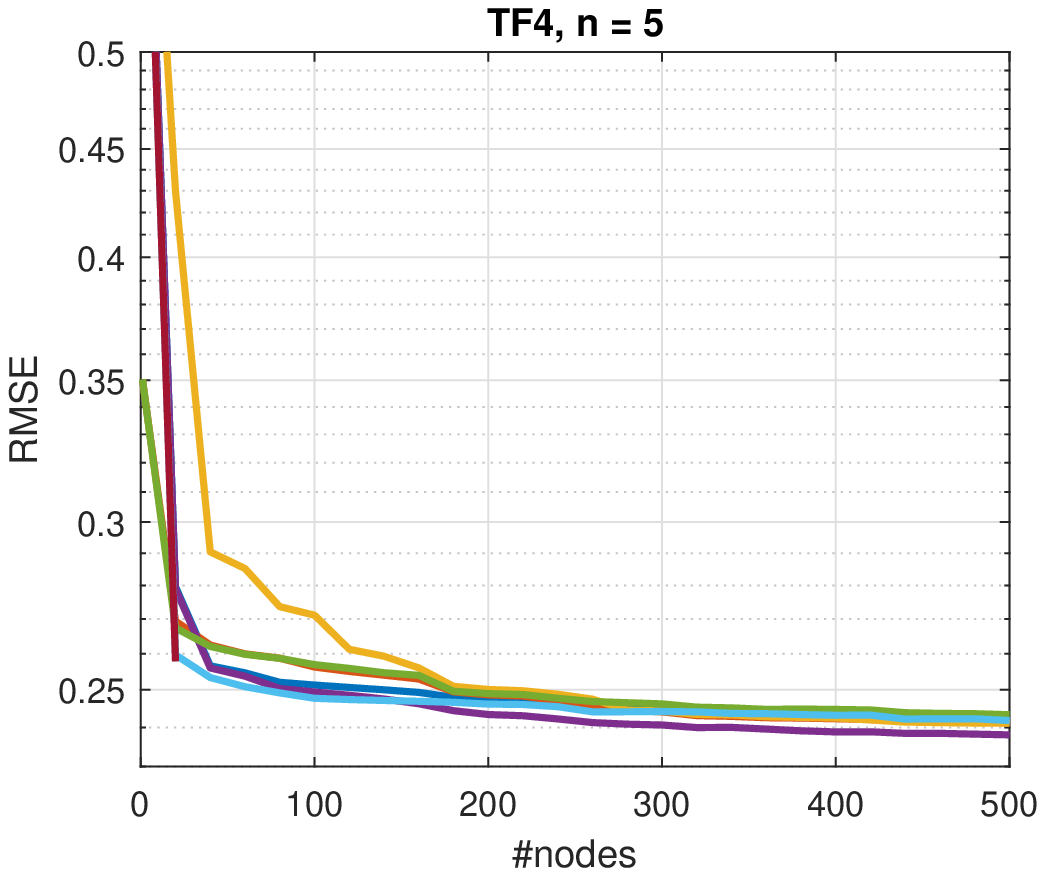}
\includegraphics[width=0.32\textwidth]{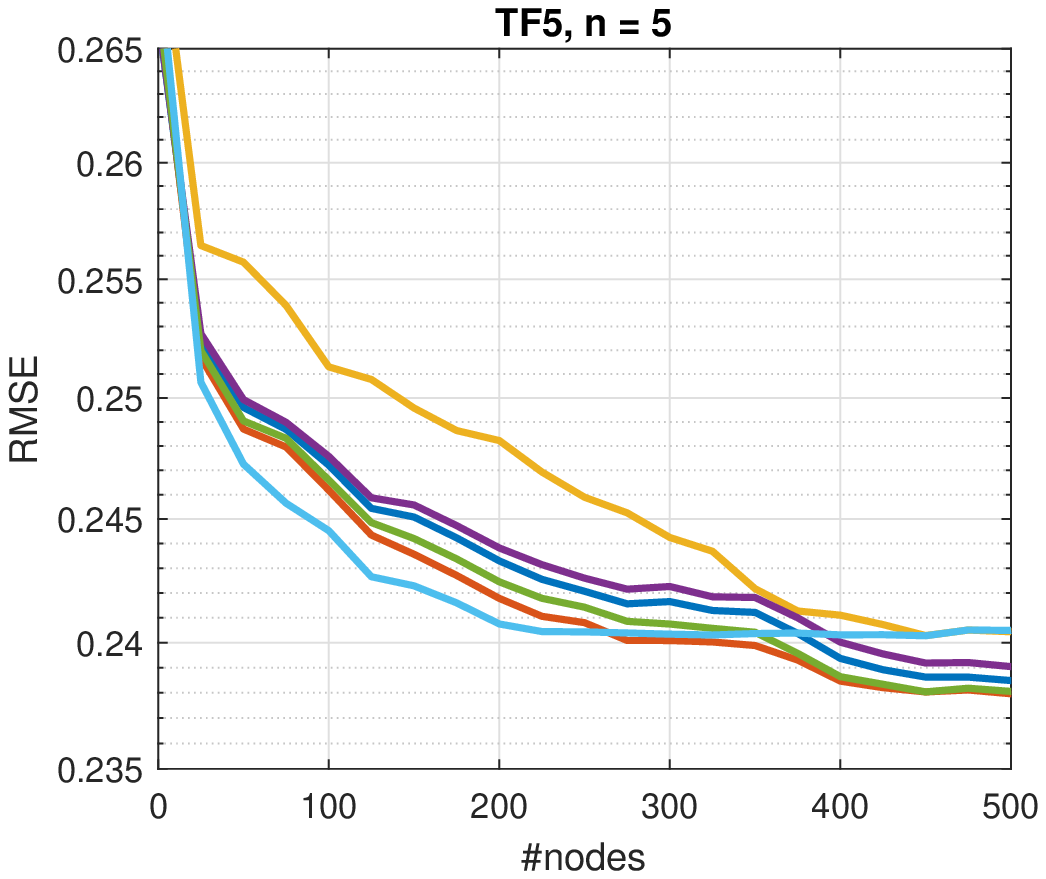}
\caption{Convergence of FNN for TF3-TF5, $n=5$.} \label{figZbn5}
\end{figure}

\begin{figure}
\centering
\includegraphics[width=0.32\textwidth]{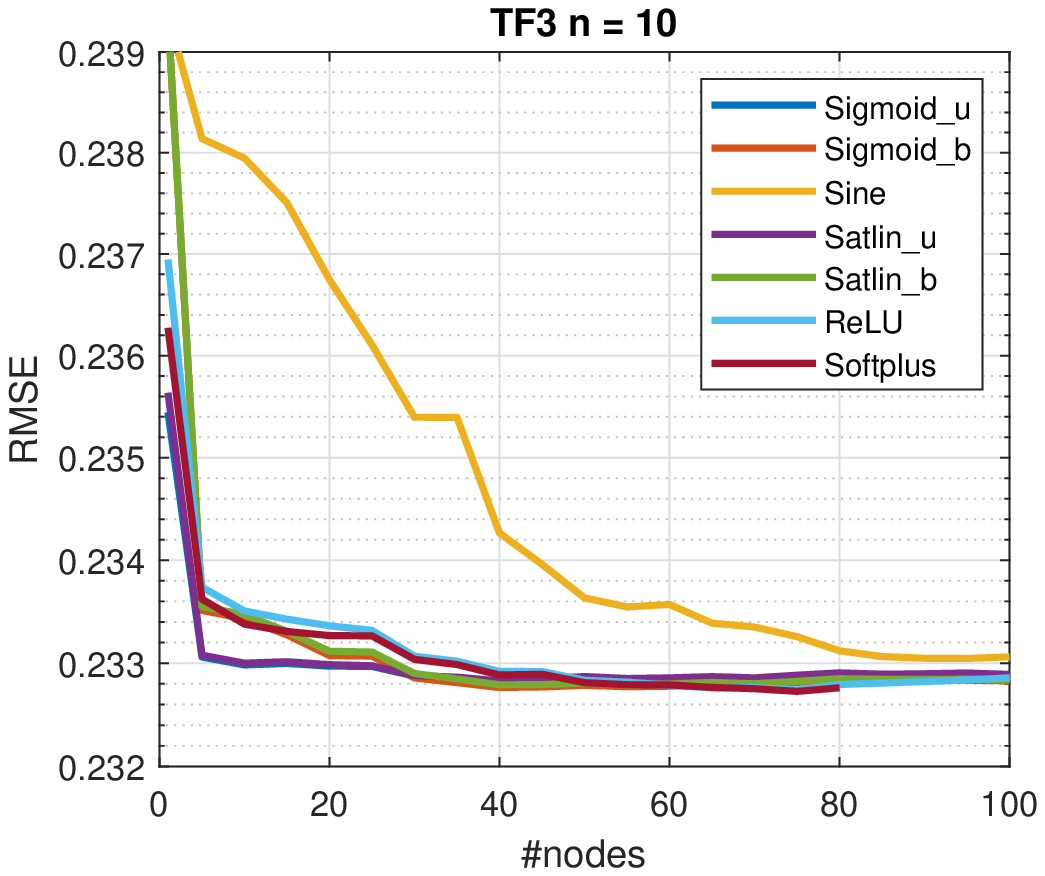}
\includegraphics[width=0.32\textwidth]{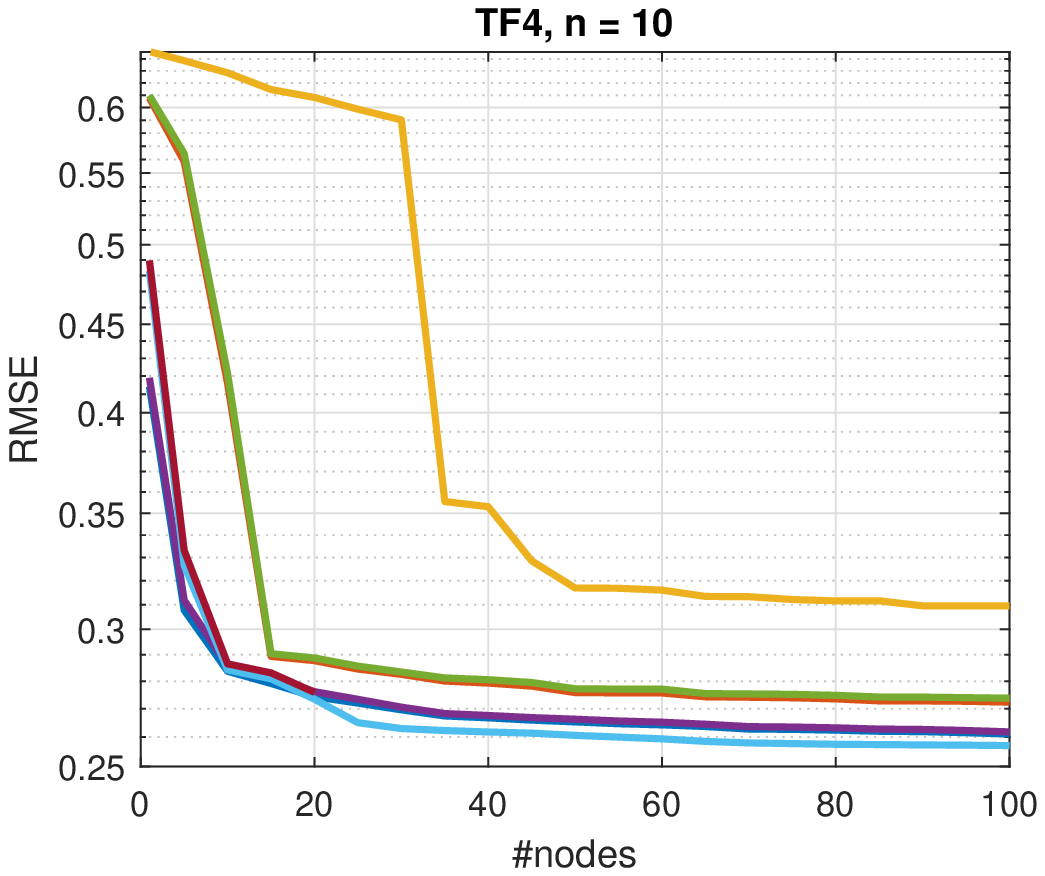}
\includegraphics[width=0.32\textwidth]{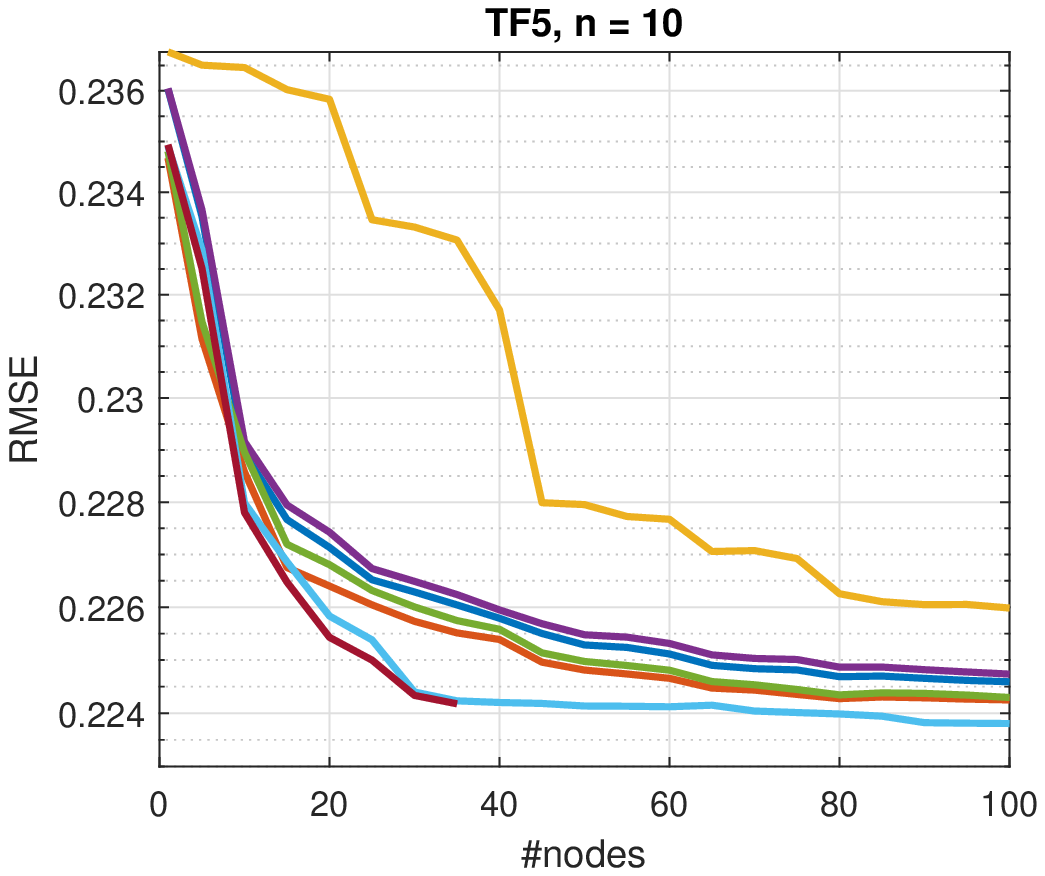}
\caption{Convergence of FNN for TF3-TF5, $n=10$.} \label{figZbn10}
\end{figure}

In the case of multidimensional modeling ($n=5$ and 10), results for all AFs were comparable (see Figs. \ref{figZbn5} and \ref{figZbn10}; $k=n$). This could be explained by the change in the TF landscape, which flattens with an increasing number of dimensions. When modeling flat TF, the AF shape turned out not to be as important as in the case of TF with strong fluctuations.

It is obvious from the performed simulations that the approximation properties of FNN trained using D-DM strongly depend on the AF type. The most useful for smoothing highly nonlinear TFs with fluctuations turned out to be the sigmoid AFs. The piecewise linear functions, i.e. $\relu$, $\satu$, and $\satb$, have problems with modeling smoothly complex TFs. Their linear parts do not fit accurately to TF nonlinearities. Likewise $\sine$ AFs cannot build an acceptable fitted function for the fluctuated TFs. The reason for this is probably the periodic nature of $\sine$. When $\sine$ AF is introduced into the input space to improve the fitted function in region $\Psi(\mathbf{x}^*)$, it can worsen the fitted function in other regions by introducing unwanted fluctuations. $\soft$ AF gave slightly worse results than sigmoid AFs for one-argument TFs, but it caused numerical problems for multivariate TFs. 

%So, the fluctuation amplitude is smaller and smaller.

%For flat TFs all AFs Another problem, which is common in multidimensional modelling, is data sparsity.    In case of sparsity, training points do not reflect TF, especially in the fluctuation regions. Thus, TF is approximated roughly with high error on      

%For higher dimensional fitting problems, i.e. 5- and 10-argument TF, the FNN converged to the similar error level for all AF types, except $\soft$ which caused numerical problems (see Figs. \ref{figZbn5}, \ref{figZbn10} and Table \ref{tab2}). In all these cases, the convergence of FNN with $\sine$ AFs was the slowest.  

%Fig. \ref{figZb12}     
%Note that the training and test points express the TFs without noise. This is because in this study we focus on investigation of the fitting properties of D-DM, not its generalization properties.  

\section{Conclusion}

The data-driven FNN learning described in this study is an alternative to both standard gradient-based learning and randomized learning. It allows us to bypass the tedious iterative process of tuning weights based on gradients. In the proposed approach, the parameters of hidden nodes are calculated based on the local properties of the TF. The AFs, which compose the fitted function, are introduced into the input space in randomly selected regions and their slopes are adjusted to the TF slopes in these regions. Consequently, the set of AFs reflects the TF fluctuations in different regions, which leads to accurate approximation. Our approach is completely different from typical randomized learning, where the AF parameters are chosen randomly and do not reflect the TF landscape. D-DM finds the network parameters quickly, without repeatedly presenting the training set.

FNN performance strongly depends on AF shape. In this work, using a data-driven approach, we derived equations for the hidden node parameters for different AFs. As our experimental study has shown, the best FNN performance in smoothing highly nonlinear TFs was achieved by the sigmoid AFs. They were able to fit to the TF fluctuations. $\relu$ AF, which is very popular in deep learning, fared very poorly in fluctuation modeling due to its piecewise linear nature. Its smooth counterpart, $\soft$, produced much better results but suffered from numerical problems related to rapid growth.


\end{document}